%% file: colm2026_conference.tex
\documentclass{article} 
\usepackage[final]{colm2026_conference}

\usepackage{microtype}
\usepackage{hyperref}
\usepackage{etoc}
\usepackage{url}
\usepackage{booktabs}

\usepackage{lineno}

\usepackage[T1]{fontenc}
\usepackage{graphicx}
\usepackage{amsmath}
\usepackage{amssymb}
\usepackage{multirow}
\usepackage{subcaption}
\usepackage[ruled,lined,linesnumbered]{algorithm2e}
\SetAlgoNlRelativeSize{-1}
\DeclareFontSeriesDefault[sf]{bf}{bx}
\usepackage[capitalize,noabbrev]{cleveref}

\usepackage{pgfplots}
\usepgfplotslibrary{groupplots}
\pgfplotsset{compat=1.18}
\usepgfplotslibrary{fillbetween}
\usepackage{tikz}
\usetikzlibrary{positioning, arrows.meta, shapes.geometric, decorations.pathreplacing, calc}
\usetikzlibrary{shadows.blur, backgrounds}

\usepackage{float}
\usepackage{longtable}
\usepackage{colortbl}
\usepackage{listings}

\definecolor{jacobcolor}{RGB}{31,119,180}
\definecolor{aswincolor}{RGB}{214,39,40}
\definecolor{zhikuncolor}{RGB}{44,160,44}
\definecolor{bencolor}{RGB}{148,103,189}

\definecolor{bestcol}{RGB}{0,120,80}
\definecolor{oursrow}{RGB}{235,241,250}
\newcommand{\best}[1]{\textcolor{bestcol}{\textbf{#1}}}
\definecolor{codebg}{RGB}{248,248,248}
\definecolor{codekw}{RGB}{0,112,32}
\definecolor{codestr}{RGB}{186,33,33}
\definecolor{codecomment}{RGB}{96,96,96}
\lstdefinestyle{pythonstyle}{
  language=Python,
  basicstyle=\ttfamily\scriptsize,
  keywordstyle=\color{codekw}\bfseries,
  stringstyle=\color{codestr},
  commentstyle=\color{codecomment}\itshape,
  backgroundcolor=\color{codebg},
  frame=single,
  framesep=4pt,
  xleftmargin=2pt,
  xrightmargin=2pt,
  breaklines=true,
  columns=fullflexible,
  keepspaces=true,
  showstringspaces=false,
  tabsize=2,
  aboveskip=0.5em,
  belowskip=0.3em,
}

\usepackage[most]{tcolorbox}

\definecolor{promptbg}{RGB}{245,247,250}
\definecolor{promptframe}{RGB}{100,120,160}
\definecolor{solverbg}{RGB}{245,250,245}
\definecolor{solverframe}{RGB}{80,140,90}
\tcbset{
  promptbox/.style={
    enhanced,
    breakable,
    colback=promptbg,
    colframe=promptframe,
    fonttitle=\bfseries\sffamily,
    boxrule=0.6pt,
    arc=2pt,
    left=8pt, right=8pt, top=6pt, bottom=6pt,
    toptitle=3pt, bottomtitle=3pt,
    attach boxed title to top left={yshift=-2mm, xshift=4mm},
    boxed title style={colback=promptframe, colframe=promptframe, arc=1.5pt, boxrule=0pt},
    fontupper=\small\ttfamily,
  },
  solverbox/.style={
    enhanced,
    breakable,
    colback=solverbg,
    colframe=solverframe,
    fonttitle=\bfseries\sffamily,
    boxrule=0.6pt,
    arc=2pt,
    left=8pt, right=8pt, top=6pt, bottom=6pt,
    toptitle=3pt, bottomtitle=3pt,
    attach boxed title to top left={yshift=-2mm, xshift=4mm},
    boxed title style={colback=solverframe, colframe=solverframe, arc=1.5pt, boxrule=0pt},
    fontupper=\small\ttfamily,
  }
}

\definecolor{vd85col}{RGB}{76,114,176}
\definecolor{vd75col}{RGB}{221,132,82}
\newcommand{\legendsquare}[1]{\protect\fcolorbox{#1}{#1}{\rule{0pt}{0.5em}\rule{0.5em}{0pt}}}
\definecolor{darkblue}{rgb}{0, 0, 0.5}
\hypersetup{colorlinks=true, citecolor=darkblue, linkcolor=darkblue, urlcolor=darkblue,
  pdftitle={Vocabulary Dropout for Curriculum Diversity in LLM Co-Evolution},
  pdfauthor={Jacob Dineen, Aswin RRV, Zhikun Xu, Ben Zhou}}

\title{Vocabulary Dropout for Curriculum Diversity\\ in LLM Co-Evolution}

  \author{Jacob Dineen, Aswin RRV, Zhikun Xu, Ben Zhou \\
  Arizona State University \\
  \texttt{\{jdineen, aravik13, zhikunxu, xzhou202\}@asu.edu}
  }

\newcommand{\tss}[1]{{\scriptsize\,$\pm$\,#1}}

\begin{document}

\ifcolmsubmission
\linenumbers
\fi

\maketitle

\etocsettocdepth.toc{none}

\begin{abstract} Co-evolutionary self-play, where one language model generates problems and another solves them, promises curriculum learning without human supervision. The promise breaks down early in practice. The proposer converges to a narrow distribution of problems that satisfy the reward function, and the collapsed curriculum teaches the solver little, stalling the loop. We introduce vocabulary dropout, a lightweight intervention that randomly masks the proposer's output logits during both policy training and curriculum generation. The mask is hard and non-stationary, so the proposer cannot lock into fixed token sequences. Training Qwen3-4B and Qwen3-8B on mathematical reasoning via R-Zero, vocabulary dropout sustains proposer diversity throughout training across lexical, semantic, and functional measures, and improves the solver by an average of +4.4 points at 8B with the largest gains on competition-level benchmarks. Explicit action-space constraints, filling the structural role that game rules fill in classical self-play, can keep co-evolution in language productive. Vocabulary dropout is one simple way to impose them.\footnote{Code available at \url{https://github.com/ARC-ASU/vocab_dropout}.} \end{abstract}

\section{Introduction}
\label{sec:introduction}

Self-play is a promising training paradigm in which models improve through competition or interaction with copies of themselves. This approach has been particularly successful in games, where the environment provides strong structural constraints on learning. In domains such as Go, StarCraft, and Dota, the rules of the game specify legal actions, fixed dynamics, and explicit objectives, making self-play well-defined~\citep{tesauro1995temporal}. Beyond these constraints, methods such as population-based training, opponent sampling, and fictitious self-play~\citep{vinyals2019alphastar,openai2019dota,heinrich2016deep} expose agents to a broader mixture of opponents and help reduce cycling or collapse to narrow strategies.

Recent co-evolutionary training frameworks apply a similar idea to language by splitting a single model into two roles, a \emph{proposer} that generates reasoning problems and a \emph{solver} that attempts them~\citep{huang2025rzero}. The proposer is trained to produce problems at the edge of the solver's ability, and the solver is trained on the resulting curriculum, creating a loop analogous to two-player self-play training. As the solver improves, the proposer gradually adapts, generating increasingly difficult problems to maintain the training signal. In practice, however, the proposer converges to a narrow set of question templates that satisfy the reward function, and the curriculum loses its value as a training signal~\citep{liu2026selfplay,chae2025towards,pu2026survivecollapseasymmetricroles}. Our experiments show that this stagnation sets in early in self-play training, coinciding with the early plateau in solver capability reported in prior work~\citep{huang2025rzero}. Standard diversity metrics fail to detect it.

We hypothesize that this collapse stems from a difference between games and language. Game environments are symbolically verifiable, so constructing a hard position generally requires finding one that is strategically demanding, and an opponent cannot stay challenging without actually improving. Natural-language questions offer no comparable verification. Neither the proposer nor the solver is bound by external rules here, and no environment certifies on its own that a question is difficult. This leaves the proposer able to satisfy the reward function without any grasp of why a question is hard. It settles on narrow templates that reliably induce solver errors or disagreement, and the curriculum ends up superficially varied while resting on a small set of underlying question templates.

Prior work attempts to mitigate this diversity collapse by anchoring the co-evolutionary loop to external signals~\citep{zhao2025absolutezero,huang2025rzero,liu2025spice,wilf2025propose}, and \citet{liu2026selfplay} argue more broadly that self-play stalls when the generated data ceases to provide learnable structure. These approaches improve stability but none directly constrain the proposer's output space.

AlphaZero injects exploration noise scaled to the number of legal moves to prevent degenerate play~\citep{silver2017mastering}, OpenAI Five randomizes game properties to force strategic diversity~\citep{openai2019dota}, and more broadly the structure of the action space shapes what policies can learn~\citep{chandak2019learning,farquhar2020growing,sharma2026ollmoptionsbasedlargelanguage}. We introduce \textbf{vocabulary dropout} (VD) to fill a structurally similar role in language generation. Where game rules constrain semantically meaningful actions, VD works one level down, on the token action space, as a hard non-stationary mask over the proposer's output logits that removes a random subset of tokens each batch. A blocked token has probability zero rather than merely reduced probability, so optimization pressure cannot erode the constraint. And since the mask is redrawn every batch, the proposer can route around any particular masked token but has no way to pre-commit to a fixed substitution strategy (\S\ref{sec:vocab_constraints}).

We evaluate vocabulary dropout within R-Zero~\citep{huang2025rzero}, which rewards the proposer through solver self-consistency rather than an external verifier, training Qwen3-4B and Qwen3-8B base models via GRPO~\citep{shao2024deepseekmath}. Our contributions are threefold. We characterize proposer diversity dynamics across co-evolution iterations empirically, showing that functional stagnation sets in early and goes unnoticed by standard lexical metrics. We introduce vocabulary dropout as a simple hard constraint on the proposer's action space, and show that it sustains diversity across lexical, semantic, and functional measures for the whole run. And we show that the sustained diversity improves the solver, by +4.4 points on average at 8B, with the largest gains on competition-level benchmarks.

\section{Related work}
\label{sec:related}

\paragraph{Self-play and co-evolution for LLMs.}
A growing family of methods trains LLMs on curricula they generate themselves~\citep{rrv2026midtrainingselfgenerateddataimproves}. SPIN refines an SFT model by having it distinguish human text from its own~\citep{chen2024selfplayfinetuningconvertsweak}. Absolute Zero, R-Zero, MAE, and Socratic-Zero mine increasingly challenging tasks from minimal seed data, through executable environments or co-evolving challenger-solver pairs~\citep{zhao2025absolutezero,huang2025rzero,chen2025mae,socratic_zero,bailey2026scalingselfplayselfguidance}. R-Few adds lightweight human supervision to stabilize self-evolution~\citep{r_few}, and SPICE grounds generation in retrieved documents~\citep{liu2025spice}. Closest in spirit, \citet{mishra2026preventing} counter curriculum collapse with a semantic coverage signal over an embedding-induced partition, though this too intervenes on problem selection and leaves the proposer's action space untouched. We run our experiments in R-Zero's two-model setup, where the challenger is directly observable and diversity interventions can be studied under control.

\paragraph{Mode collapse in LLM training.}
Policy-gradient methods concentrate probability on high-reward behaviors, causing entropy collapse even when rewards are verifiable~\citep{zhou2025breaking,gai2025differentialsmoothingmitigatessharpening}. \citet{cui2025entropy} establish that downstream performance is bottlenecked by entropy exhaustion. This applies not only to solvers but also to proposers in co-evolutionary settings, where the proposer's policy gradient training drives it toward a narrow set of high-reward problem templates. RL-trained reasoning models exhibit reduced solution diversity~\citep{yue2025reinforcement,karouzos2026where} and may fail to develop reasoning behaviors absent from the base policy~\citep{rrv2025thinktuning}, while LLM populations grow homogeneous over generations~\citep{jiang2025artificialhivemindopenendedhomogeneity,chae2025towards}. Proposed mitigations generally target the reward or sampling side, including entropy regularization~\citep{cui2025entropy}, structured reward decomposition~\citep{dineen2025qa,ye2025cclearn}, diversity-promoting rewards~\citep{li2025jointlyreinforcingdiversityquality}, population-based training~\citep{jaderberg2017population}, and inference-time noise injection~\citep{khalid2026noise}. Vocabulary dropout instead acts directly on the action space rather than the reward or sampling distribution.

\paragraph{Action-space design in RL.} In classical RL, action representations shape exploration and generalization~\citep{chandak2019learning,farquhar2020growing}. In games, rule-imposed constraints on legal moves are what make self-play productive~\citep{silver2017mastering,openai2019dota}. For language models, the vocabulary \emph{is} the action space, and its size is typically fixed at the tokenizer's full output. Vocabulary reduction has been studied for efficiency~\citep{nozaki2025vocab}, and RLPT~\citep{pang2026reinforcement} masks contextually irrelevant tokens to concentrate the policy on promising outputs. Our vocabulary dropout inverts this logic, masking tokens randomly and non-stationarily to prevent concentration rather than encourage it.

\section{Preliminaries: GRPO and R-Zero}
\label{sec:preliminaries}

\begin{figure*}[t]
\centering
\includegraphics[width=\textwidth]{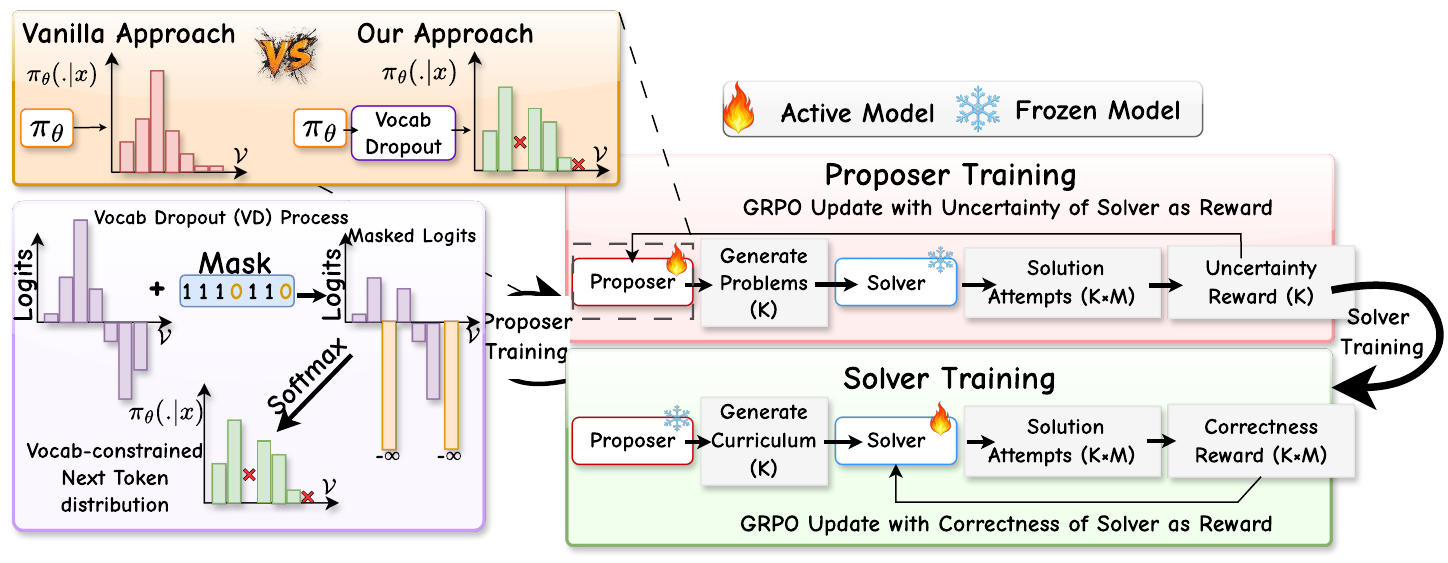}
\caption{Training pipeline. \textbf{Left}: Vocabulary dropout masks a random subset of output logits, constraining the proposer's token distribution. Orange denotes masked tokens. \textbf{Right}: The co-evolution loop. In \textbf{Phase 1} (proposer training), the proposer generates $K$ problems, the frozen solver attempts each $M$ times, and the proposer is rewarded based on solver uncertainty. In \textbf{Phase 2} (solver training), the frozen proposer generates a curriculum of $K$ problems and the solver is rewarded for matching the correct answer.}
\label{fig:pipeline}
\end{figure*}

R-Zero~\citep{huang2025rzero} trains a proposer $\pi_P$ and a solver $\pi_S$ for mathematical reasoning, both from one base checkpoint, alternating between the two roles (\Cref{fig:pipeline}, right). The optimizer in both phases is GRPO~\citep{shao2024deepseekmath}. GRPO avoids a learned value function altogether. It samples $G$ responses $\{o_1, \ldots, o_G\}$ from $\pi_\theta$ for a prompt $x$, gives each a reward $r_i$, and treats the group statistics as the baseline, $\hat{A}_i = (r_i - \mu_G) / \sigma_G$. The policy is then updated by maximizing the clipped surrogate:
\begin{equation}
\mathcal{J}_{\text{GRPO}}(\theta) = \mathbb{E}\left[\frac{1}{G}\sum_{i=1}^{G} \min\!\left(\rho_i \hat{A}_i,\; \text{clip}(\rho_i, 1{-}\epsilon, 1{+}\epsilon)\hat{A}_i\right) - \beta\, D_{\text{KL}}(\pi_\theta \| \pi_{\text{ref}})\right]
\end{equation}
with $\rho_i = \pi_\theta(o_i|x) / \pi_{\text{old}}(o_i|x)$ the importance ratio and a KL penalty holding $\pi_\theta$ near the reference model.

R-Zero rewards the proposer for problems that sit near the edge of what the solver can currently do. Each generated problem $q$ goes to the frozen solver, which answers it $M$ times, and the reward comes from how consistent those answers are with one another. Near-unanimous agreement signals a problem too easy to be useful. Heavy disagreement signals one that is too hard, or simply ill posed. Reward is highest at maximal solver uncertainty:
\begin{equation}
  r_P(q) = \begin{cases}
    \min(\text{acc}(q), 1 - \text{acc}(q)) & \text{if } \text{acc}(q) \in [\tau_{\min}, \tau_{\max}] \\
    0 & \text{otherwise}
  \end{cases}
\end{equation}
where $\text{acc}(q)$ is the fraction of solver responses agreeing with the majority vote. R-Zero additionally penalizes intra-batch repetition via pairwise BLEU similarity, clustering near-duplicate questions and subtracting a penalty proportional to cluster size from the uncertainty reward, which encourages surface-level diversity within each batch. The solver trains on problems filtered from the proposer's output, receiving binary reward for matching the majority-vote pseudo-label. 

\section{Method}
\label{sec:method}

\subsection{Vocabulary dropout}
\label{sec:vocab_constraints} 
Given the full vocabulary $\mathcal{V}$ with a small protected subset $\mathcal{F} \subset \mathcal{V}$ of format-critical tokens exempt from masking, such as the question tags and boxed-answer syntax (details in \S\ref{app:vocab_mask}), we define a retention probability $\alpha \in (0, 1]$. At each batch $b$, we sample a mask $\mathbf{m}^{(b)} \in \{0,1\}^{|\mathcal{V}|}$ where each entry is drawn independently: 

\begin{equation} m_v^{(b)} \sim \begin{cases} \text{Bernoulli}(\alpha) & \text{if } v \notin \mathcal{F} \\ 1 & \text{if } v \in \mathcal{F} \end{cases} \end{equation} 

The masked logits at batch $b$ are then: 

\begin{equation} \tilde{\ell}_v^{(b)} = \begin{cases} \ell_v & \text{if } m_v^{(b)} = 1 \\ -\infty & \text{otherwise} \end{cases} \end{equation} 

where $\ell_v$ is the original logit for token $v$. The resulting token distribution $\tilde{\pi}^{(b)}(\cdot | x) = \text{softmax}(\tilde{\boldsymbol{\ell}}^{(b)})$ varies across batches even for identical inputs, since the surviving token set $\mathcal{V}^{(b)} = \{v : m_v^{(b)} = 1\}$ is resampled every batch. 

The mask is applied in both phases of each iteration. During GRPO training, the mask is applied at the sampling stage of each rollout, so the policy can only generate from unmasked tokens. The subsequent policy-gradient update computes log-probabilities over the full vocabulary, allowing gradients to flow through all tokens. Because the available tokens change every batch, the policy cannot concentrate probability on a fixed set of high-reward sequences, which slows entropy collapse~\citep{cui2025entropy} and preserves exploration across the token distribution~\citep{wang2025beyond}. During curriculum generation, the per-batch mask likewise forces the proposer to produce varied phrasings and problem structures. The proposer therefore uses a larger effective vocabulary despite fewer tokens per batch, as confirmed in \S\ref{sec:results_diversity}. Implementation details are in \S\ref{app:vocab_mask}.

\subsection{Training procedure}
\label{sec:training}

Both models are trained with GRPO~\citep{shao2024deepseekmath} in alternating phases (pseudocode in \Cref{alg:coevolution}, prompt templates in \S\ref{app:prompts}). The proposer generates competition-level math problems via GRPO under vocabulary dropout with retention probability $\alpha$. Each rollout is checked for valid format, and valid problems are scored by the frozen solver using the self-consistency reward from \S\ref{sec:preliminaries}. After the proposer finishes training, it generates a large question set with vocabulary dropout still active. Problems within the difficulty window ($[\tau_{\min}, \tau_{\max}]{=}[0.3, 0.7]$, following R-Zero) are retained, and the solver trains on this filtered set via GRPO with a binary reward for matching the proposer's stated answer. This differs from R-Zero, which uses the solver's majority-vote pseudo-label as the reward target.

\section{Experiments}
\label{sec:experiments}

\subsection{Setup}

\paragraph{Vocabulary dropout.}
We apply vocabulary dropout in both phases (GRPO training and curriculum generation) and test retention probabilities $\alpha \in \{0.75, 0.85\}$ at each model scale, with $\alpha = 1.0$, the full vocabulary, as the baseline. Removing 15\% or 25\% of the non-protected vocabulary brackets the range where the tradeoff between diversity and output coherence shows up most clearly. Train-only and gen-only ablations isolate where the mask matters most.

\paragraph{Models and training.}
Base models are Qwen3-4B-Base and Qwen3-8B-Base~\citep{yang2025qwen3}, with the proposer and solver both initialized from the same checkpoint as in R-Zero~\citep{huang2025rzero}. Each run takes 2 NVIDIA H200 GPUs and goes for 5 co-evolution iterations of GRPO~\citep{shao2024deepseekmath} on \texttt{verl}~\citep{verl} with \texttt{vLLM}~\citep{kwon2023vllm} serving rollouts, longer than R-Zero's 3-iteration protocol because peak evaluation performance did not always land on the final iteration there. Proposer GRPO samples $G{=}4$ responses per prompt at temperature 1.0 with a global batch size of 16, and proposer scoring queries the frozen solver $M{=}10$ times per question for self-consistency. The solver then trains on the filtered curriculum at global batch size 8. Optimization uses AdamW, learning rate $1{\times}10^{-6}$, weight decay $10^{-2}$, and a KL penalty of $\beta{=}10^{-2}$, with full hyperparameters in \S\ref{app:hyperparameters}.

We also train the two entropy-control methods of \citet{cui2025entropy}, KL-Cov and Clip-Cov, which target the highest-covariance tokens in the proposer's policy update and are applied to the proposer only (\S\ref{app:klcov}). Raising the proposer's sampling temperature, the simplest intervention of all, breaks the pipeline before the first iteration completes, since at $T \geq 1.5$ the proposer cannot fill a batch with valid question-answer pairs.

\paragraph{Diversity metrics.} We track proposer diversity at three granularities. Self-BLEU covers the lexical level, catching surface repetition. The semantic level uses two metrics, the Vendi score~\citep{friedman2023vendi}, an effective count of distinct question types in embedding space, and novelty rate, the fraction of questions whose nearest-neighbor cosine distance to all prior iterations exceeds 0.3, both on \texttt{text-embedding-3-small}~\citep{openai2024embeddings} embeddings. For the functional level we compute epiplexity, which \citet{finzi2026entropy} introduced and \citet{liu2026selfplay} applied to self-play curricula, measuring the learnable information content of generated questions through prequential minimum description length. Mean difficulty, proposer policy entropy ($H_\pi$), question length, and unique token counts are reported alongside (\S\S\ref{sec:results_diversity},~\ref{app:vocab_utilization}).

\paragraph{Standard benchmarks.}
Solver capability is evaluated on GSM8K~\citep{cobbe2021gsm8k}, MATH~\citep{hendrycks2021math}, AMC~\citep{amc2023}, OlympiadBench~\citep{he2024olympiadbench}, and AIME 2024/2025~\citep{aime2024}, which together span grade-school arithmetic up to olympiad problems. Annealing experiments (\S\ref{sec:discussion}) add Minerva Math~\citep{lewkowycz2022minerva}. The untrained base models are evaluated under the same protocol as a reference point. Evaluation is zero-shot with the model's default chat template (\S\ref{app:eval_prompt}), temperature 0.7, and a maximum context of 4096 tokens, scored by exact match via \texttt{math\_verify}~\citep{huggingface_math_verify_2025}. Every number is an average over 3 independent runs with mean $\pm$ standard error~\citep{miller2024adding}.

\section{Results}
\label{sec:results}
\input{tables/vanilla_eval}

\subsection{Downstream performance}

\Cref{tab:vanilla_eval} reports pass@1 accuracy at the final co-evolution iteration. We avoid selecting the best checkpoint or run post hoc to prevent selection bias~\citep{cawley2010overfitting,dodge2019show}. Vocabulary dropout improves solver performance at both scales. We use VD$\langle\alpha\rangle$ as shorthand for vocabulary dropout at retention probability $\alpha$ (e.g., VD75 means $\alpha{=}0.75$). At 8B, VD75 leads all configurations with a +4.4 point average improvement over the baseline, driven by strong gains on competition-level tasks (AMC, Olympiad, AIME). At 4B, VD85 improves over the baseline on average (39.3 vs.\ 38.3). All configurations, including the baseline, improve over the untrained base models. Both entropy-control methods improve over the no-regularization baseline at 8B (41.6 and 41.5 vs.\ 39.4 avg) but fall short of the best VD configuration at each scale. At 4B, KL-Cov falls below the baseline (37.0 vs.\ 38.3) while Clip-Cov roughly matches it (38.7).

On the proposer side, co-evolutionary training does not improve downstream reasoning ability (\Cref{tab:proposer_eval}). The 8B proposer (VD75: 38.4, baseline: 38.8) slightly degrades relative to the base model (38.9), and 4B shows a similar pattern. At 4B, proposer performance drops modestly under dropout (VD85: 31.9 vs.\ baseline: 32.8), consistent with the tighter constraint on the smaller model. The proposer is optimized for calibrated problem generation rather than problem solving, and these objectives do not appear to transfer.

\paragraph{Ablations.} \Cref{tab:vanilla_eval} also isolates the effects of masking strength and dropout phase. At 4B, VD85 outperforms the baseline while VD75 underperforms, so the optimal masking strength is scale-dependent and an overly aggressive mask can exceed a small model's capacity to compensate. The phase decomposition at 4B shows gen-only slightly edging train-only (37.8 vs.\ 37.4), but neither recovers the baseline, because what binds at this scale is masking strength rather than dropout phase.

At 8B the pattern flips. Stronger masking consistently improves performance, both single-phase variants surpass the baseline (42.2 and 41.8 vs.\ 39.4), and the model absorbs the full mask with no loss in output quality. Gen-only edges train-only here too, and the combination outperforms either alone. What separates the scales is whether the model has enough capacity to absorb VD75 at all. Embedding-level analysis of these phase ablations (\S\ref{app:embedding_ablations}, \Cref{tab:question_diversity}) confirms that gen-only contributes the larger share of diversity, but the combination sustains the strongest growth across iterations.

\subsection{Proposer diversity}
\label{sec:results_diversity}

\paragraph{Baseline stagnation is early and hidden.}
Nearly all of the baseline proposer's functional collapse happens between iterations 1 and 2, after which the metrics plateau (\Cref{fig:diversity_combined}). In a single iteration, mean difficulty (panel c) jumps from 0.50 to 0.86 at 4B and from 0.71 to 0.92 at 8B, past the $\tau_{\max}{=}0.7$ threshold on solver accuracy. Policy entropy (a), Vendi score (d), and epiplexity (f) lock in just as early. Self-BLEU and novelty rate (b, e), on the other hand, stay flat from start to finish, so the proposer is recycling similar questions in a way surface metrics simply miss. Diversity interventions applied after the first iteration may accordingly be too late to head off curriculum collapse.

\paragraph{Dropout sustains diversity across all tiers.}
Under dropout, all functional metrics continue to evolve past iteration 2, the proposer's distribution remains responsive to the solver's improvement rather than freezing after the first update, and every diversity metric we measure improves (\Cref{fig:diversity_combined}). At the \textit{lexical} level, Self-BLEU is roughly $2{\times}$ lower under dropout at both scales (b). At the \textit{semantic} level, the Vendi score shows a gap of ${\sim}15$ effective question types at both scales (d), mean pairwise cosine distance with Qwen3-Embedding-4B~\citep{li2025jointlyreinforcingdiversityquality} is $+4$--$6\%$ higher under dropout, and NLI-based logical diversity~\citep{stasaski-hearst-2022-semantic} is $+19$--$68\%$ higher (\Cref{tab:nli_diversity}). Novelty rate is ${\sim}1.5$--$2{\times}$ higher under dropout (e). At the \textit{functional} level, epiplexity is ${\sim}35\%$ higher under dropout (f), indicating substantially more learnable structure in the curriculum (computation details in \S\S\ref{app:vendi},~\ref{app:epiplexity}). These rankings are robust to an encoder swap and unidirectional NLI scoring (\S\ref{app:embedding_diversity}). Cumulative Vendi scores (\S\ref{app:embedding_diversity}) show the same pattern, with the baseline plateauing after iteration 2--3 while VD75 continues to grow, holding ${\sim}35\%$ higher cumulative diversity at both scales.

\Cref{fig:question_complexity} confirms this at the structural level. Dropout questions contain more distinct numeric values and are substantially longer, while baseline runs produce roughly $2{\times}$ more valid questions per iteration, consistent with simpler problems that pass format and reward filters at a higher rate. Representative question pairs in \S\ref{app:qualitative} illustrate the contrast. At the token level, VD75 proposers use 36--52\% more unique tokens than the baseline (3{,}697 vs.\ 2{,}717 at 4B, 2{,}868 vs.\ 1{,}881 at 8B) despite fewer being available at any given step (\S\ref{app:vocab_utilization}). The non-stationary mask forces broader vocabulary use rather than concentration. KL-Cov and Clip-Cov stay within 2\% of baseline on all four metrics at both scales, leaving the structural profile unchanged where vocabulary dropout reshapes it.

\input{figures/question_profile}

\input{figures/diversity_combined}

\subsection{Curriculum quality in isolation}
\label{sec:curriculum_transfer}

The solver gains in \Cref{tab:vanilla_eval} arise inside the co-evolution loop, which couples curriculum content with self-consistency reward noise and the live proposer-solver pairing. To test whether the gains come from the curriculum itself, we isolate it. We take the iteration-5 question sets from the baseline, KL-Cov, Clip-Cov, and VD ($\alpha{=}0.75$) 8B runs, re-verify every problem with an external model (GPT-5-mini~\citep{singh2025openai}), keep only those whose independently-derived answer matches the curriculum's, roughly 20--31\% of each arm, and GRPO-train a fresh Qwen3-4B-Base on each verified subset (training details in \S\ref{app:curriculum_transfer}).

\begin{table}[H]
\centering
\caption{Solver pass@1 (\%) after GRPO on each arm's externally-verified iteration-5 curriculum, training a fresh Qwen3-4B-Base. Curricula are taken from the 8B co-evolution runs and verified with GPT-5-mini. Best trained arm per column in \best{green bold}.}
\label{tab:curriculum_transfer}
\footnotesize
\setlength{\tabcolsep}{3.5pt}
\begin{tabular}{l ccccccc c}
\toprule
\textbf{Trained on} & MATH500 & GSM8K & AMC & Olympiad & AIME'24 & AIME'25 & Minerva & \textbf{Avg.} \\
\midrule
\textcolor{gray}{Base (no training)} & \textcolor{gray}{55.0} & \textcolor{gray}{75.1} & \textcolor{gray}{25.0} & \textcolor{gray}{20.0} & \textcolor{gray}{6.7} & \textcolor{gray}{3.3} & \textcolor{gray}{25.4} & \textcolor{gray}{30.1} \\
Baseline & \best{71.0} & 88.6 & 40.0 & \best{28.2} & 3.3 & 6.7 & 29.8 & 38.2 \\
KL-Cov & 69.8 & 88.6 & 45.0 & 27.6 & \best{10.0} & 3.3 & 31.6 & 39.4 \\
Clip-Cov & 66.6 & 86.7 & 40.0 & 24.2 & \best{10.0} & 3.3 & 32.0 & 37.5 \\
\rowcolor{oursrow} VD & \best{71.0} & \best{89.8} & \best{50.0} & 27.8 & 3.3 & \best{10.0} & \best{37.1} & \best{41.3} \\
\bottomrule
\end{tabular}
\end{table}

All four trained arms improve substantially over the untrained base, so the comparison ranks curricula that each carry real signal. VD's curriculum produces the strongest student of the four, beating the baseline curriculum by $+3.1$ on average and the next-best arm (KL-Cov) by $+1.9$. Clip-Cov's curriculum falls below the baseline, so once verification noise is stripped its content is worse than R-Zero's self-judged data. VD's advantage holds even with externally verified answers and a fresh model, so it comes from the curriculum it produces rather than the self-consistency reward.

\paragraph{Diversity is not an artifact of ill-posedness.}
Generation-phase dropout lowers the proposer's answer correctness by roughly $7\%$ at convergence (\Cref{tab:correctness}), which raises the question of whether the added diversity simply reflects more ill-posed problems. To check, we isolate the well-posed portion of each curriculum at the final iteration, having GPT-4.1-mini label every problem in the trained difficulty band as well-posed or ill-posed and solve it independently (details in \S\ref{app:wellposed}).

\begin{table}[H]
\centering
\caption{Curriculum composition and diversity at the final iteration. \emph{Filter pass} and \emph{Pass rate} count generated problems landing in the difficulty band $[0.3, 0.7]$, \emph{Well-posed} and \emph{Correct} are GPT-4.1-mini judgments, and $\Delta$ Vendi is VD75 minus baseline on the well-posed (WP) and correct (C) subsets.}
\label{tab:wellposed}
\small
\setlength{\tabcolsep}{5pt}
\begin{tabular}{ll cccc cc}
\toprule
\textbf{Scale} & \textbf{Method} & Filter pass & Pass rate & Well-posed & Correct & $\Delta$ Vendi (WP) & $\Delta$ Vendi (C) \\
\midrule
\multirow{2}{*}{4B} & Baseline & 946 & 11.8\% & 850 & 275 & --- & --- \\
& \cellcolor{oursrow} VD75 & \cellcolor{oursrow}1{,}222 & \cellcolor{oursrow}15.3\% & \cellcolor{oursrow}925 & \cellcolor{oursrow}269 & \cellcolor{oursrow}$+11.4$ & \cellcolor{oursrow}$+7.1$ \\
\midrule
\multirow{2}{*}{8B} & Baseline & 510 & 6.4\% & 465 & 163 & --- & --- \\
& \cellcolor{oursrow} VD75 & \cellcolor{oursrow}1{,}408 & \cellcolor{oursrow}17.6\% & \cellcolor{oursrow}1{,}112 & \cellcolor{oursrow}325 & \cellcolor{oursrow}$+7.5$ & \cellcolor{oursrow}$+2.8$ \\
\bottomrule
\end{tabular}
\end{table}

Vocabulary dropout produces more usable training data (\Cref{tab:wellposed}). At 8B the baseline lands only 6.4\% of its generated problems in the difficulty band, against VD75's 17.6\%, and VD75 yields roughly twice as many oracle-correct problems per iteration (325 vs.\ 163). The diversity gain also survives the correctness control. VD75's Vendi exceeds the baseline by $+11.4$ at 4B and $+7.5$ at 8B on the well-posed subset, NLI logical diversity favors VD75 on the same subset (\S\ref{app:wellposed}), and both gaps persist on the strictly-correct problems, so the added diversity reflects a more varied well-posed curriculum rather than extra noise.

\subsection{Generality beyond math and co-evolution}
\label{sec:generality}

\paragraph{Out-of-domain transfer.}
The results so far measure mathematical reasoning inside the R-Zero loop, so we also evaluate the 8B VD75 solver zero-shot on BBH~\citep{suzgun2022bbh}, MedQA-USMLE~\citep{jin2021medqa}, SuperGPQA~\citep{du2025supergpqa}, and WinoGrande~\citep{sakaguchi2021winogrande}, none of which appear in training. The solver beats the R-Zero baseline on three of the four, by $+10.1$ on BBH, $+5.0$ on MedQA-USMLE, and $+2.1$ on SuperGPQA, trailing by $4.5$ on WinoGrande (\Cref{tab:ood}). Vocabulary dropout constrains the proposer's output distribution rather than encoding anything math-specific, and the gains from its curriculum carry to domains the training set never covers.

\begin{table}[H]
\centering
\caption{\textbf{Left}: zero-shot out-of-domain transfer of the 8B solver (VD at $\alpha{=}0.75$), evaluated under the protocol of \Cref{tab:vanilla_eval}. Best between Baseline and VD per row in \best{green bold}. \textbf{Right}: policy mean per-token entropy over the first 100 GRPO steps of single-policy training (Qwen3-4B-Base on GSM8K), baseline vs.\ VD ($\alpha{=}0.95$).}
\label{tab:ood}\label{tab:vanilla_entropy}
\footnotesize
\setlength{\tabcolsep}{3pt}
\begin{minipage}[t]{0.5\textwidth}
\vspace{0pt}
\centering
\begin{tabular}{l ccc c}
\toprule
\textbf{Benchmark} & \textcolor{gray}{Base} & Baseline & VD & $\Delta$ \\
\midrule
BBH         & \textcolor{gray}{37.5} & 44.8 & \best{55.0} & $+10.1$ \\
MedQA-USMLE & \textcolor{gray}{41.2} & 60.8 & \best{65.8} & $+5.0$ \\
SuperGPQA   & \textcolor{gray}{27.0} & 27.7 & \best{29.8} & $+2.1$ \\
WinoGrande  & \textcolor{gray}{66.1} & \best{69.1} & 64.6 & $-4.5$ \\
\bottomrule
\end{tabular}
\end{minipage}\hfill
\begin{minipage}[t]{0.48\textwidth}
\vspace{0pt}
\centering
\begin{tikzpicture}
\definecolor{entbase}{RGB}{120,120,120}
\definecolor{entvd}{RGB}{221,132,82}
\begin{axis}[
  width=\linewidth, height=3.0cm,
  xmin=0, xmax=99, ymin=0, ymax=2.05,
  xlabel={GRPO step}, ylabel={Entropy},
  xlabel style={font=\scriptsize, yshift=3pt}, ylabel style={font=\scriptsize, yshift=-6pt},
  ytick={0,1,2},
  tick label style={font=\scriptsize},
  axis line style={gray!50}, tick style={gray!30},
  legend style={font=\scriptsize, draw=gray!40, rounded corners=1pt, fill=white, at={(0.96,0.92)}, anchor=north east, inner xsep=3pt, inner ysep=2pt},
  legend cell align=left,
  every axis plot/.append style={line width=0.9pt},
]
\addplot[entbase, dashed] coordinates {(0,0.5857)(1,1.2702)(2,0.9366)(3,1.0526)(4,1.3787)(5,0.4314)(6,0.6217)(7,1.0707)(8,0.6092)(9,0.5612)(10,0.8443)(11,0.4332)(12,0.3466)(13,0.2752)(14,0.3826)(15,0.2934)(16,0.3169)(17,0.6720)(18,0.2252)(19,0.2292)(20,0.2810)(21,0.4506)(22,0.3941)(23,0.5475)(24,0.1983)(25,0.2119)(26,0.2419)(27,0.2797)(28,0.1691)(29,0.2244)(30,0.2689)(31,0.1769)(32,0.3014)(33,0.1996)(34,0.2434)(35,0.1653)(36,0.2087)(37,0.1955)(38,0.1425)(39,0.2288)(40,0.1696)(41,0.1603)(42,0.1641)(43,0.1373)(44,0.1597)(45,0.1381)(46,0.1273)(47,0.2038)(48,0.1332)(49,0.1154)(50,0.1703)(51,0.1752)(52,0.1629)(53,0.1875)(54,0.1574)(55,0.1224)(56,0.1656)(57,0.1642)(58,0.1278)(59,0.1520)(60,0.1266)(61,0.1070)(62,0.1675)(63,0.1253)(64,0.1102)(65,0.2034)(66,0.1950)(67,0.1200)(68,0.1632)(69,0.0969)(70,0.1118)(71,0.1429)(72,0.1165)(73,0.0973)(74,0.2225)(75,0.2448)(76,0.1281)(77,0.0723)(78,0.2563)(79,0.0923)(80,0.1052)(81,0.1066)(82,0.0918)(83,0.1138)(84,0.1411)(85,0.0985)(86,0.1327)(87,0.0968)(88,0.1283)(89,0.1192)(90,0.0744)(91,0.2047)(92,0.1052)(93,0.1031)(94,0.0769)(95,0.1054)(96,0.1159)(97,0.1394)(98,0.1217)(99,0.1048)};
\addlegendentry{Baseline ($\alpha{=}1.0$)}
\addplot[entvd] coordinates {(0,1.9800)(1,1.3812)(2,1.3057)(3,0.7891)(4,1.1583)(5,0.8999)(6,1.0630)(7,0.9870)(8,1.5879)(9,0.7470)(10,1.0132)(11,0.9815)(12,1.3982)(13,0.7676)(14,0.6912)(15,0.4472)(16,0.6757)(17,0.5490)(18,0.4725)(19,0.7670)(20,0.5924)(21,0.7194)(22,1.0169)(23,0.5881)(24,0.6776)(25,0.4575)(26,0.3497)(27,0.8583)(28,0.6628)(29,0.4050)(30,0.4438)(31,0.4554)(32,0.4622)(33,0.2926)(34,0.4217)(35,0.3785)(36,0.3411)(37,0.7552)(38,0.3547)(39,0.6099)(40,0.3298)(41,0.2874)(42,0.3661)(43,0.3226)(44,0.3789)(45,0.2668)(46,0.2782)(47,0.3625)(48,0.3281)(49,0.3751)(50,0.3294)(51,0.3080)(52,0.3490)(53,0.3828)(54,0.2854)(55,0.2513)(56,0.3473)(57,0.2979)(58,0.4178)(59,0.3837)(60,0.3404)(61,0.3131)(62,0.3719)(63,0.3700)(64,0.4730)(65,0.3812)(66,0.3189)(67,0.3238)(68,0.3107)(69,0.2865)(70,0.4809)(71,0.3093)(72,0.3008)(73,0.3210)(74,0.3170)(75,0.3608)(76,0.3639)(77,0.3775)(78,0.3959)(79,0.3612)(80,0.3241)(81,0.4352)(82,0.3348)(83,0.3863)(84,0.3122)(85,0.4475)(86,0.4110)(87,0.4166)(88,0.3067)(89,0.3972)(90,0.3882)(91,0.3765)(92,0.2629)(93,0.3622)(94,0.4654)(95,0.3578)(96,0.3955)(97,0.3422)(98,0.4364)(99,0.3774)};
\addlegendentry{VD ($\alpha{=}0.95$)}
\end{axis}
\end{tikzpicture}
\end{minipage}
\end{table}

\paragraph{A single policy without self-play.}
To probe the mechanism outside co-evolution entirely, we train Qwen3-4B-Base on GSM8K with standard GRPO and a binary correctness reward, with and without a light mask ($\alpha{=}0.95$) on the policy's rollouts (setup and per-benchmark results in \S\ref{app:vanilla_grpo}). Accuracy stays close, with dropout ahead on four of seven benchmarks, and the clear effect is on entropy (\Cref{tab:vanilla_entropy}). The unregularized policy collapses from 0.59 to 0.10, the failure mode documented by \citet{cui2025entropy}, while vocabulary dropout holds entropy roughly $2\times$ higher on average and $3.8\times$ higher at convergence. Even without a co-evolving partner, the mask acts as an exploration regularizer, the same mechanism we argue sustains proposer diversity in the loop.

\section{Discussion}
\label{sec:discussion}

\paragraph{Action-space constraints for language self-play.} Our results support the hypothesis that explicit action-space constraints sustain productive co-evolution in language. R-Zero already includes a reward-side repetition penalty based on pairwise BLEU, and the baseline proposer still collapses to a narrow distribution within two iterations, so reward-side diversity pressure alone appears insufficient. Vocabulary dropout constrains the output space directly instead, and because the mask moves every batch there is no fixed template distribution for the proposer to settle into. The diversity gains (\S\ref{sec:results_diversity}), the solver improvements (\Cref{tab:vanilla_eval}), and the slower entropy collapse outside co-evolution (\S\ref{sec:generality}) are all consistent with this explanation.

\paragraph{Curriculum quality vs.\ curriculum diversity.}
Proposer correctness degrades in every condition as training progresses, from $56$--$81\%$ at the first iteration to roughly $32$--$39\%$ by the last (\Cref{tab:correctness}). The degradation reflects a limit of self-consistency verification that the baseline and vocabulary dropout share, namely that solver disagreement cannot separate hard problems from ill-posed ones. Diversity collapse is the orthogonal problem, and that is the one vocabulary dropout addresses. The baseline produces a narrow, repetitive curriculum of declining quality, while dropout maintains a varied one whose well-posed subset is rich enough to offset the noise. \citet{shao2025spurious} report that GRPO can amplify pre-training priors even under spurious rewards, and \citet{setlur2024rl} find that incorrect synthetic data can still help when negative signals are well structured, both consistent with what we observe. Stronger verification layered on top could keep the diversity while filtering the noise (\S\ref{app:illposed}).

\paragraph{Non-stationarity and scale.}
The masking-strength ablations (\Cref{tab:vanilla_eval}) show that vocabulary dropout's value is scale-dependent, a capacity effect. Larger models carry more redundant token representations and can reroute around masked tokens, so a more aggressive mask raises diversity without sacrificing coherence. This points toward adaptive masks that scale $\alpha$ with model size or anneal it over training. A linear anneal from $\alpha{=}0.75$ to $1.0$ supports this reading, peaking two iterations earlier than fixed VD75, winning three of five iterations on math average, and tracing a more stable trajectory, with an OlympiadBench range of 3.0 points against 6.4 for fixed masking (\S\ref{app:annealing}). It also attains the highest cumulative performance across the five iterations.

\paragraph{Scope and limitations.}
When the proposer already has more capacity than the solver (8B$\to$4B cross-scale, \S\ref{app:cross_scale}), dropout hurts rather than helps ($-1.4$ avg), because the difficulty calibration is already misaligned and diversity pressure amplifies the mismatch. Even without dropout, the cross-scale pairing does no better than the symmetric baseline (38.8 vs.\ 38.3 avg), so a stronger proposer alone does not buy a better curriculum. The matched-scale 8B curriculum still trains a 4B model well in static transfer (\S\ref{sec:curriculum_transfer}), so the live difficulty calibration rather than the curriculum content drives this degradation. Similarly, co-evolution with or without dropout does not consistently benefit instruction-tuned models in our Qwen2.5-1.5B-Instruct experiments (\S\ref{app:instruct}). 
These boundaries track the co-evolutionary setting rather than vocabulary dropout itself. R-Zero-style self-play is most productive when proposer and solver are capacity-matched base models, the range in which vocabulary dropout operates.

\section{Conclusion}
\label{sec:conclusion}

We showed that a simple, non-stationary constraint on the proposer's output vocabulary sustains curriculum diversity and improves solver performance in co-evolutionary self-play. Like dropout on hidden units, which prevents co-adaptation of neurons, it softens co-adaptation of token sequences. The method needs no auxiliary models or reward modifications, keeps the curriculum from collapsing onto trivially easy problems, and its gains carry beyond math at 8B (\S\ref{sec:generality}). Our results suggest that action-space structure, rather than reward design alone, is a productive lever for co-evolutionary dynamics in language, and composing the mask with stronger verification is a natural route to self-play curricula that stay both varied and reliable.

\section*{Ethics Statement}
This work studies training dynamics in co-evolutionary LLM systems using mathematical reasoning as a testbed. The models, data, and methods are standard in the field. We do not foresee specific negative societal consequences beyond those common to language model research.

\section*{Reproducibility Statement}
All of our code is released at \url{https://github.com/ARC-ASU/vocab_dropout}, with runbooks covering the vocabulary dropout runs, the R-Zero baseline, and the KL-Cov and Clip-Cov arms. The same repository holds the diversity metric implementations and pins the dependency versions we ran with. Hyperparameters, prompt templates, and the verification procedure appear in \S\ref{app:hyperparameters}, \S\ref{app:prompts}, and \S\ref{app:verification}. Reported numbers are means over 3 seeds with standard errors.

\bibliography{colm2026_conference}
\bibliographystyle{colm2026_conference}

\clearpage
\appendix

\etocsettocdepth.toc{subsection}
\etocsettocstyle{\section*{Appendix Contents}}{}
\tableofcontents
\clearpage

\section{Algorithm}
\label{app:algorithm}

\Cref{alg:coevolution} gives the full co-evolutionary training loop.

\begin{algorithm}[H]
\caption{Vocabulary Dropout Co-Evolutionary Self-Play}\label{alg:coevolution}
\DontPrintSemicolon
\SetKwInOut{Input}{Input}
\SetKwInOut{Output}{Output}
\SetKwFunction{BernoulliMask}{BernoulliMask}
\Input{Base model $\theta_0$;\; retention prob.\ $\alpha$;\; difficulty window $[\tau_{\min}, \tau_{\max}]$;\; iterations $T$;\; proposer GRPO group size $G$;\; solver self-consistency samples $M$}
\Output{Trained proposer $\pi_P^{(T)}$ and solver $\pi_S^{(T)}$}
\BlankLine
$\pi_P^{(0)},\; \pi_S^{(0)} \gets \theta_0$\;
$\mathcal{F} \gets$ protected format-critical tokens \tcp*{always survive}
\BlankLine
\For{$t = 1, \ldots, T$}{
  \tcc{\textbf{Phase 1: Train proposer} (solver frozen)}
  \For{each GRPO batch}{
    $\mathcal{M}^{(b)} \gets \{v \in \mathcal{V} : m_v \sim \text{Bern}(\alpha)\} \cup \mathcal{F}$ \tcp*{fresh mask per batch}
    Sample $\{o_i\}_{i=1}^{G}$ from $\pi_P$, setting logits to $-\infty$ for tokens $\notin \mathcal{M}^{(b)}$\;
    \For{each rollout $o_i$}{
      Parse $q_i, a_i$ from $o_i$ \tcp*{question, boxed answer}
      \lIf{parse fails}{$r_i^P \gets 0$;\; \textbf{continue}}
      Query frozen $\pi_S^{(t-1)}$ with $q_i$;\; collect $M$ responses\;
      $\text{acc}_i \gets$ fraction matching majority vote\;
      $r_i^P \gets \min(\text{acc}_i,\; 1{-}\text{acc}_i) \cdot \mathbf{1}[\text{acc}_i \in [\tau_{\min}, \tau_{\max}]]$\;
    }
    Update $\pi_P$ via GRPO on $\{(o_i, r_i^P)\}$\;
  }
  $\pi_P^{(t)} \gets \pi_P$\;
  \BlankLine
  \tcc{\textbf{Phase 2: Generate curriculum, then train solver}}
  $\mathcal{Q} \gets$ sample problems from $\pi_P^{(t)}$ with per-batch $\mathcal{M}^{(b)}$\;
  $\mathcal{Q}_{\text{train}} \gets \{(q, a) \in \mathcal{Q} : \text{acc}(q) \in [\tau_{\min}, \tau_{\max}]\}$ \tcp*{$a$ is the proposer's stated answer}
  \For{each GRPO batch}{
    Sample solver rollouts for $(q, a) \in \mathcal{Q}_{\text{train}}$\;
    $r^S \gets \mathbf{1}[\text{solver answer} = a]$\;
    Update $\pi_S$ via GRPO on $\{(q, r^S)\}$\;
  }
  $\pi_S^{(t)} \gets \pi_S$\;
}
\Return $\pi_P^{(T)},\; \pi_S^{(T)}$\;
\end{algorithm}

\section{Implementation details}
\label{app:implementation}

\subsection{Verification via solver self-consistency}
\label{app:verification}

Following R-Zero~\citep{huang2025rzero}, we verify proposer outputs using the solver itself rather than an external reward model. Each proposer output is checked for valid format (\texttt{<question>} tags and a \texttt{\textbackslash boxed\{\}} answer). Valid questions are sent to the frozen solver, which attempts each problem 10 times. The proposer reward is $r = \min(\text{acc}, 1{-}\text{acc})$, where $\text{acc}$ is the fraction of solver responses matching the majority vote. This peaks at 50\% solve rate, incentivizing problems at the solver's decision boundary.

For solver training, generated questions are filtered to the difficulty window $[\tau_{\min}, \tau_{\max}] = [0.3, 0.7]$, following R-Zero.

\subsection{Vocabulary dropout implementation}
\label{app:vocab_mask}

Vocabulary dropout rides on vLLM's \texttt{allowed\_token\_ids} parameter in \texttt{SamplingParams}, which sets logits to $-\infty$ for every token outside the allowed set before sampling. At the start of each batch we draw a fresh Bernoulli mask over the full vocabulary, add the protected set $\mathcal{F}$ of format-critical tokens needed for structural validity and answer formatting, and hand the resulting token ID list to vLLM. Nothing in the model weights or the sampling kernel changes. \Cref{fig:code_diff} shows the code change relative to the standard R-Zero rollout.

\definecolor{diffgreen}{RGB}{230,255,230}
\definecolor{diffred}{RGB}{255,230,230}
\definecolor{diffgreentext}{RGB}{0,100,0}
\definecolor{diffredtext}{RGB}{160,0,0}
\definecolor{diffgray}{RGB}{100,100,100}

\begin{figure}[H]
\centering
\begin{lstlisting}[
  basicstyle=\ttfamily\scriptsize,
  frame=leftline,
  framerule=0.6pt,
  rulecolor=\color{gray},
  framesep=4pt,
  xleftmargin=2pt,
  xrightmargin=2pt,
  breaklines=true,
  columns=fullflexible,
  keepspaces=true,
  showstringspaces=false,
  tabsize=2,
  aboveskip=0.5em,
  belowskip=0.3em,
  backgroundcolor=\color{white},
  moredelim={**[is][\color{diffredtext}\bfseries]{@@r}{@@}},
  moredelim={**[is][\color{diffgreentext}\bfseries]{@@g}{@@}},
  moredelim={**[is][\color{diffgray}]{@@d}{@@}},
  escapechar=|,
]
 def rollout(model, prompts,
|\colorbox{diffred}{\makebox[\dimexpr\linewidth-20pt][l]{\texttt{\textcolor{diffredtext}{-\phantom{xx}sampling\_params):}}}}|
|\colorbox{diffgreen}{\makebox[\dimexpr\linewidth-20pt][l]{\texttt{\textcolor{diffgreentext}{+\phantom{xx}sampling\_params,}}}}|
|\colorbox{diffgreen}{\makebox[\dimexpr\linewidth-20pt][l]{\texttt{\textcolor{diffgreentext}{+\phantom{xx}tokenizer, alpha=0.75):}}}}|
|\colorbox{diffgreen}{\makebox[\dimexpr\linewidth-20pt][l]{\texttt{\textcolor{diffgreentext}{+\phantom{x}V = tokenizer.vocab\_size}}}}|
|\colorbox{diffgreen}{\makebox[\dimexpr\linewidth-20pt][l]{\texttt{\textcolor{diffgreentext}{+\phantom{x}mask = torch.bernoulli(}}}}|
|\colorbox{diffgreen}{\makebox[\dimexpr\linewidth-20pt][l]{\texttt{\textcolor{diffgreentext}{+\phantom{xxxxx}torch.full((V,), alpha)}}}}|
|\colorbox{diffgreen}{\makebox[\dimexpr\linewidth-20pt][l]{\texttt{\textcolor{diffgreentext}{+\phantom{x}).bool()}}}}|
|\colorbox{diffgreen}{\makebox[\dimexpr\linewidth-20pt][l]{\texttt{\textcolor{diffgreentext}{+\phantom{x}\# Protect format-critical tokens}}}}|
|\colorbox{diffgreen}{\makebox[\dimexpr\linewidth-20pt][l]{\texttt{\textcolor{diffgreentext}{+\phantom{x}for t in get\_protected\_token\_ids(}}}}|
|\colorbox{diffgreen}{\makebox[\dimexpr\linewidth-20pt][l]{\texttt{\textcolor{diffgreentext}{+\phantom{xxxxx}tokenizer):}}}}|
|\colorbox{diffgreen}{\makebox[\dimexpr\linewidth-20pt][l]{\texttt{\textcolor{diffgreentext}{+\phantom{xxx}mask[t] = True}}}}|
|\colorbox{diffgreen}{\makebox[\dimexpr\linewidth-20pt][l]{\texttt{\textcolor{diffgreentext}{+\phantom{x}sampling\_params.allowed\_token\_ids = (}}}}|
|\colorbox{diffgreen}{\makebox[\dimexpr\linewidth-20pt][l]{\texttt{\textcolor{diffgreentext}{+\phantom{xxxxx}mask.nonzero(as\_tuple=True)[0]}}}}|
|\colorbox{diffgreen}{\makebox[\dimexpr\linewidth-20pt][l]{\texttt{\textcolor{diffgreentext}{+\phantom{xxxxx}.tolist())}}}}|
   completions = model.generate(
       prompts, sampling_params
   )
   return completions
\end{lstlisting}
\caption{Vocabulary dropout as a unified diff. The only change is sampling a Bernoulli mask and passing the surviving token IDs to vLLM's \texttt{allowed\_token\_ids} before generation. The mask is resampled every batch.}
\label{fig:code_diff}
\end{figure}

\subsection{Compute requirements}
\label{app:compute}

Every run uses 2 NVIDIA H200 GPUs. In the proposer phase one GPU handles GRPO training and the other hosts a vLLM server for self-consistency scoring, while curriculum generation and question evaluation split across both. The solver phase trains on both GPUs. Each experiment covers 5 iterations and 3 independent seeds per configuration, and benchmark evaluation runs on 4--8 NVIDIA H100 GPUs through vLLM.

Vocabulary dropout itself adds no compute beyond the standard rollout, since the mask is enforced at the logit level inside the existing sampling kernel.

\subsection{Entropy-control baselines}
\label{app:klcov}

We compare against the two entropy-control methods of \citet{cui2025entropy}, which prevent policy entropy collapse by limiting updates on tokens whose log-probability and advantage co-vary most strongly. KL-Cov adds a per-token KL penalty to the top-$k$ such tokens (their Eq.~14), while Clip-Cov detaches the gradient on a random subset of tokens whose covariance falls in $[\omega_\text{low}, \omega_\text{high}]$, leaving the standard PPO loss elsewhere. Both are applied to the proposer only, matching VD's scope, with all other training settings identical to the Baseline and VD arms. We list the hyperparameters in \Cref{tab:klcov_hypers}.

We also tried the simplest diversity intervention, raising the proposer's sampling temperature. At $T \geq 1.5$ the proposer degrades enough that it cannot populate a full batch of valid question-answer pairs, so R-Zero's filtering and extraction break in the first co-evolution iteration. We therefore do not report higher temperature as a baseline.

\begin{table}[H]
\centering
\caption{Entropy-control hyperparameters, following \citet{cui2025entropy}'s 7B reference configuration.}
\label{tab:klcov_hypers}
\small
\begin{tabular}{lll}
\toprule
\textbf{Method} & \textbf{Hyperparameter} & \textbf{Value} \\
\midrule
KL-Cov & $k$ (top-covariance token fraction) & $2 \times 10^{-3}$ \\
 & $\beta$ (KL coefficient on selected tokens) & 1.0 \\
\midrule
Clip-Cov & $\rho$ (masked-token fraction) & $2 \times 10^{-4}$ \\
 & $\omega_\text{low}, \omega_\text{high}$ (covariance bounds) & 1.0, 5.0 \\
\midrule
Both & Entropy bonus coefficient & 0 \\
 & Scope & Proposer only \\
\bottomrule
\end{tabular}
\end{table}

\section{Qualitative examples}
\label{app:qualitative}

We present representative problems generated by the baseline and vocabulary dropout proposers at the same co-evolution iteration, as well as examples of ill-posed problems that both proposers produce.

\subsection{Baseline vs.\ vocabulary dropout}

The following pairs are drawn from iteration-5 question sets of the 8B proposer. For each pair, the baseline and dropout examples share the same broad semantic theme. Dropout examples were filtered to those with solver accuracy $\geq 0.7$.

\noindent
\begin{minipage}[t]{0.48\columnwidth}
\begin{tcolorbox}[enhanced, equal height group=qual1,
  colback=gray!5, colframe=gray!60,
  fonttitle=\bfseries\scriptsize, title={Baseline 8B -- Baking},
  boxrule=0.4pt, arc=1pt, left=4pt, right=4pt, top=3pt, bottom=3pt,
  fontupper=\scriptsize]
\textit{Sarah is baking cookies for her class. She has 5 trays, and each tray can hold 12 cookies. If she wants to bake 60 cookies in total, how many batches does she need to bake if each batch uses all 5 trays?}
\end{tcolorbox}
\end{minipage}%
\hfill
\begin{minipage}[t]{0.48\columnwidth}
\begin{tcolorbox}[enhanced, equal height group=qual1,
  colback=oursrow, colframe=promptframe!60,
  fonttitle=\bfseries\scriptsize, title={VD75 8B -- Baking},
  boxrule=0.4pt, arc=1pt, left=4pt, right=4pt, top=3pt, bottom=3pt,
  fontupper=\scriptsize]
\textit{Sarah is baking cookies to sell at the bake sale to raise funds for the kindergarten program at her local elementary school. She plans to bake three types: sugar cookies, oat cakes, and granola bars. She baked 120 cookies overall. If half are sugar cookies and four-fifths of the remaining are oat cakes, how many granola bars did she bake?}
\end{tcolorbox}
\end{minipage}

\vspace{0.6em}
\noindent
\begin{minipage}[t]{0.48\columnwidth}
\begin{tcolorbox}[enhanced, equal height group=qual2,
  colback=gray!5, colframe=gray!60,
  fonttitle=\bfseries\scriptsize, title={Baseline 8B -- Farming},
  boxrule=0.4pt, arc=1pt, left=4pt, right=4pt, top=3pt, bottom=3pt,
  fontupper=\scriptsize]
\textit{A farmer has a rectangular field that is 30 meters long and 20 meters wide. He wants to plant apple trees in rows spaced 5 meters apart and plants 3 trees in each row. How many apple trees can the farmer plant in his field?}
\end{tcolorbox}
\end{minipage}%
\hfill
\begin{minipage}[t]{0.48\columnwidth}
\begin{tcolorbox}[enhanced, equal height group=qual2,
  colback=oursrow, colframe=promptframe!60,
  fonttitle=\bfseries\scriptsize, title={VD75 8B -- Farming},
  boxrule=0.4pt, arc=1pt, left=4pt, right=4pt, top=3pt, bottom=3pt,
  fontupper=\scriptsize]
\textit{At the local farmer stand, apples are sold at \$2 per kilo and bananas at \$3 per kilo. A customer bought 1 kilo of apples and 1 kilo of bananas on Monday, and 2 kilos of apples and 2 kilos of bananas on Tuesday. What was the customer's total expenditure over the two days?}
\end{tcolorbox}
\end{minipage}

\subsection{Ill-posed problems}
\label{app:illposed}

Self-consistency verification rewards solver disagreement, but cannot distinguish hard problems from ill-defined ones. Both the baseline and vocabulary dropout proposers produce problems where the proposer's stated ground truth is incorrect. In each case below, all 10 solvers answered correctly but were scored as wrong (score = 1.0) because they disagreed with the proposer's erroneous answer.

\newcommand{\illposedbox}[6]{%
\begin{tcolorbox}[enhanced, equal height group=#3,
  colback=#1, colframe=#2,
  fonttitle=\bfseries\scriptsize, title={#4},
  boxrule=0.4pt, arc=1pt, left=4pt, right=4pt, top=3pt, bottom=3pt,
  fontupper=\scriptsize]
\textit{#5}
\vspace{2pt}
#6
\end{tcolorbox}%
}

\vspace{0.5em}
\noindent
\begin{minipage}[t]{0.48\columnwidth}
\illposedbox{gray!5}{gray!60}{ip1}{Baseline 8B}{Sophie bought 126 seeds. She planted 34 on Saturday and 58 on Sunday. How many left?}{%
\textbf{Prop:} \textcolor{red}{92} \quad \textbf{Solvers:} \textcolor{green!50!black}{34\,\checkmark} {\tiny (10/10)}\\[1pt]
{\tiny\textit{Subtracted only Saturday; forgot Sunday. Correct: $126{-}34{-}58{=}34$.}}}
\end{minipage}%
\hfill
\begin{minipage}[t]{0.48\columnwidth}
\illposedbox{oursrow}{promptframe!60}{ip1}{VD75 8B}{Alice has 48 marigolds and 36 roses, bundles them equally (largest possible). How many marigolds per bundle?}{%
\textbf{Prop:} \textcolor{red}{48} \quad \textbf{Solvers:} \textcolor{green!50!black}{12\,\checkmark} {\tiny (10/10)}\\[1pt]
{\tiny\textit{Wrote total count (48) instead of bundle size (GCD $= 12$).}}}
\end{minipage}

\vspace{0.6em}
\noindent
\begin{minipage}[t]{0.48\columnwidth}
\illposedbox{gray!5}{gray!60}{ip2}{Baseline 8B}{It takes 30 seconds to fill a 6-liter jug. How many liters in 2 minutes?}{%
\textbf{Prop:} \textcolor{red}{4} \quad \textbf{Solvers:} \textcolor{green!50!black}{24\,\checkmark} {\tiny (10/10)}\\[1pt]
{\tiny\textit{Computed fills ($120 \div 30 = 4$) but forgot jug capacity. Correct: $4 \times 6 = 24$.}}}
\end{minipage}%
\hfill
\begin{minipage}[t]{0.48\columnwidth}
\illposedbox{oursrow}{promptframe!60}{ip2}{VD75 8B}{Mike paints a fence in 2 hr, Tom in 3 hr. How many minutes together?}{%
\textbf{Prop:} \textcolor{red}{$1\text{ min}\ldots$} \quad \textbf{Solvers:} \textcolor{green!50!black}{72\,\checkmark} {\tiny (10/10)}\\[1pt]
{\tiny\textit{Confused hourly with per-minute rates; answer truncated. Correct: 72 min.}}}
\end{minipage}

\vspace{0.6em}
\noindent
\begin{minipage}[t]{0.48\columnwidth}
\illposedbox{gray!5}{gray!60}{ip3}{Baseline 4B}{John had 50 dollars. He bought a 10 dollar ice cream. How many dollars left?}{%
\textbf{Prop:} \textcolor{red}{50} \quad \textbf{Solvers:} \textcolor{green!50!black}{40\,\checkmark} {\tiny (10/10)}\\[1pt]
{\tiny\textit{Echoed starting amount without subtracting. Correct: $50{-}10{=}40$.}}}
\end{minipage}%
\hfill
\begin{minipage}[t]{0.48\columnwidth}
\illposedbox{oursrow}{promptframe!60}{ip3}{VD75 4B}{Simon had 30 cars at 12, half of age-15 count. Simon had half of Lou's at 15. Lou's total?}{%
\textbf{Prop:} \textcolor{red}{60} \quad \textbf{Solvers:} \textcolor{green!50!black}{120\,\checkmark} {\tiny (10/10)}\\[1pt]
{\tiny\textit{Stopped one step early; attributed Simon's count to Lou. Correct: $60 \times 2 = 120$.}}}
\end{minipage}

\vspace{0.6em}
\noindent
\begin{minipage}[t]{0.48\columnwidth}
\illposedbox{gray!5}{gray!60}{ip4}{Baseline 4B}{180 pages in a book. Emma reads $\tfrac{1}{5}$ in week 1 and $\tfrac{2}{5}$ in week 2. How many in week 2?}{%
\textbf{Prop:} \textcolor{red}{36} \quad \textbf{Solvers:} \textcolor{green!50!black}{72\,\checkmark} {\tiny (10/10)}\\[1pt]
{\tiny\textit{Computed week-1 fraction instead of week-2. Correct: $\tfrac{2}{5} \times 180 = 72$.}}}
\end{minipage}%
\hfill
\begin{minipage}[t]{0.48\columnwidth}
\illposedbox{oursrow}{promptframe!60}{ip4}{VD75 4B}{John bought 4 fish, returned half. How many left?}{%
\textbf{Prop:} \textcolor{red}{4} \quad \textbf{Solvers:} \textcolor{green!50!black}{2\,\checkmark} {\tiny (10/10)}\\[1pt]
{\tiny\textit{Echoed starting count without applying return. Correct: $4{-}2{=}2$.}}}
\end{minipage}

These failures arise from a limitation of self-consistency verification, which equates solver disagreement with problem difficulty regardless of answer correctness. The proposer receives maximum reward for these problems precisely because it provided a wrong answer that no solver reproduces. Both the baseline and dropout settings exhibit this failure mode, which is a property of the verification scheme rather than vocabulary dropout.

\subsection{Proposer answer correctness}
\label{app:proposer_correctness}

To estimate the fraction of generated problems where the proposer's stated answer is correct, we use GPT-4.1-mini~\citep{openai2025gpt41} as an independent verifier. For each problem at every iteration, the verifier solves the problem independently and compares its answer to the proposer's. \Cref{tab:correctness} reports correctness trajectories across all conditions and phase ablations.

\begin{table}[H]
\centering
\caption{Proposer answer correctness (\%) over co-evolution iterations, judged by GPT-4.1-mini solving each problem independently.}
\label{tab:correctness}
\vspace{0.3em}
\small
\setlength{\tabcolsep}{5pt}
\begin{tabular}{ll ccccc}
\toprule
\textbf{Model} & \textbf{Setting} & It.\ 1 & It.\ 2 & It.\ 3 & It.\ 4 & It.\ 5 \\
\midrule
\multirow{4}{*}{Qwen3-8B}
& Baseline      & 81.4 & 45.5 & 42.2 & 39.1 & 38.0 \\
& Train-only    & 79.0 & 43.2 & 42.0 & 40.4 & 39.0 \\
& Gen-only      & 62.0 & 38.3 & 33.7 & 33.4 & 31.9 \\
& \cellcolor{oursrow} VD75 & \cellcolor{oursrow}61.8 & \cellcolor{oursrow}34.9 & \cellcolor{oursrow}33.1 & \cellcolor{oursrow}36.7 & \cellcolor{oursrow}32.4 \\
\midrule
\multirow{4}{*}{Qwen3-4B}
& Baseline      & 75.6 & 38.1 & 38.0 & 42.2 & 37.7 \\
& Train-only    & 76.0 & 37.2 & 44.2 & 38.0 & 38.2 \\
& Gen-only      & 56.0 & 34.5 & 32.1 & 29.5 & 32.4 \\
& \cellcolor{oursrow} VD75 & \cellcolor{oursrow}61.2 & \cellcolor{oursrow}31.2 & \cellcolor{oursrow}30.5 & \cellcolor{oursrow}30.7 & \cellcolor{oursrow}31.9 \\
\bottomrule
\end{tabular}
\end{table}

\begin{table}[H]
\centering
\caption{Proposer reasoning ability (pass@1 \%) on standard benchmarks. VD is applied to the proposer during co-evolution. Co-evolutionary training does not improve the proposer's own problem-solving ability.}
\label{tab:proposer_eval}
\vspace{0.3em}
\footnotesize
\setlength{\tabcolsep}{5pt}
\begin{tabular}{l cccccc c}
\toprule
& \multicolumn{6}{c}{\textbf{Pass@1 (\%)}} & \\
\cmidrule(lr){2-7}
\textbf{Setting} & MATH500 & GSM8K & AMC & Olympiad & AIME'24 & AIME'25 & \textbf{Avg.} \\
\midrule
\multicolumn{8}{c}{\textbf{Qwen3-4B}} \\
\midrule
\textcolor{gray}{Base (no training)} & \textcolor{gray}{54.1\tss{2.2}} & \textcolor{gray}{63.5\tss{5.2}} & \textcolor{gray}{29.2\tss{1.8}} & \textcolor{gray}{20.6\tss{1.3}} & \textcolor{gray}{12.2\tss{0.9}} & \textcolor{gray}{1.1\tss{0.9}} & \textcolor{gray}{30.1} \\
Baseline ($\alpha{=}1.0$) & 56.0\tss{1.1} & 72.7\tss{3.4} & 33.3\tss{3.0} & 20.3\tss{0.9} & 10.0\tss{1.6} & 4.4\tss{0.9} & 32.8 \\
\rowcolor{oursrow} VD85 ($\alpha{=}0.85$) & 58.4\tss{0.8} & 60.5\tss{4.1} & 37.5\tss{4.1} & 19.3\tss{1.1} & 11.1\tss{0.9} & 4.4\tss{2.4} & 31.9 \\
\midrule
\multicolumn{8}{c}{\textbf{Qwen3-8B}} \\
\midrule
\textcolor{gray}{Base (no training)} & \textcolor{gray}{70.7\tss{0.1}} & \textcolor{gray}{82.2\tss{0.7}} & \textcolor{gray}{45.8\tss{3.6}} & \textcolor{gray}{18.1\tss{1.3}} & \textcolor{gray}{8.9\tss{2.4}} & \textcolor{gray}{7.8\tss{0.9}} & \textcolor{gray}{38.9} \\
Baseline ($\alpha{=}1.0$) & 65.1\tss{2.1} & 87.5\tss{1.2} & 40.8\tss{3.8} & 21.8\tss{0.7} & 7.8\tss{0.9} & 10.0\tss{1.6} & 38.8 \\
\rowcolor{oursrow} VD75 ($\alpha{=}0.75$) & 61.7\tss{1.7} & 85.2\tss{1.4} & 40.0\tss{3.1} & 23.3\tss{1.6} & 11.1\tss{0.9} & 8.9\tss{1.8} & 38.4 \\
\bottomrule
\end{tabular}
\end{table}

All conditions show a sharp correctness drop from iteration 1 to 2, then stabilize. This mirrors the stagnation pattern observed in \S\ref{sec:results_diversity}, where the proposer's distribution shifts rapidly in the first iteration as it learns to target the solver's decision boundary, then locks in. Two clusters emerge at convergence. Conditions without generation-phase dropout (baseline, train-only) stabilize at ${\sim}38$--$42\%$ correctness, while conditions with generation-phase dropout (gen-only, VD75) settle at ${\sim}30$--$34\%$. The ${\sim}7\%$ gap is the price of diversity, since the non-stationary mask produces more varied but less reliably correct problems. Despite this, VD75 produces the strongest solver improvements at 8B (\Cref{tab:vanilla_eval}), suggesting that the diversity of the well-posed subset more than compensates for the increased noise. This motivates composing vocabulary dropout with stronger verification, as discussed in \S\ref{sec:discussion}.

\subsection{Curriculum diversity controlling for correctness}
\label{app:wellposed}

This section gives the methodology behind \Cref{tab:wellposed} in \S\ref{sec:curriculum_transfer}. At the final co-evolution iteration we record how many generated problems land in the productive difficulty band $[\tau_{\min}, \tau_{\max}]{=}[0.3, 0.7]$ that the solver trains on, then label each filtered problem as well-posed (its answer is determinate, whether the proposer states it correctly or not) or ill-posed, using the same GPT-4.1-mini verifier as \S\ref{app:proposer_correctness}. Beyond the Vendi gaps in \Cref{tab:wellposed}, NLI logical diversity on the well-posed subset also favors VD75, by $+0.12$ at 4B and $+0.03$ at 8B.

\section{Additional results}
\label{app:results}

\subsection{Cumulative embedding diversity}
\label{app:embedding_diversity}

\Cref{fig:diversity_combined} reports per-iteration Vendi scores and novelty rates. As a complementary view, \Cref{fig:embedding_diversity} tracks the cumulative Vendi score, where all questions from iterations 1 through $N$ are pooled (subsampled to 2{,}000), embedded with \texttt{text-embedding-3-small}~\citep{openai2024embeddings}, and scored. VD75 maintains ${\sim}35\%$ higher cumulative diversity than the baseline at both scales. The baseline plateaus after iteration 2--3, while VD75 continues to grow.

\input{figures/embedding_diversity}

As additional semantic-diversity views, \Cref{tab:nli_diversity} reports mean pairwise cosine distance with Qwen3-Embedding-4B and NLI-based logical diversity computed with DeBERTa-v3-large fine-tuned on MNLI/FEVER/ANLI/LingNLI/WANLI, on the curriculum-filtered problem sets at iteration 5. VD outperforms the baseline at every (scale, metric) cell, with relative lifts ranging from $+4$--$6\%$ (semantic) to $+19$--$68\%$ (logical). This rank order holds under an encoder swap (all-MiniLM-L6-v2) and unidirectional NLI scoring.

\begin{table}[H]
\centering
\caption{NLI and cosine diversity at iteration 5, on the curriculum-filtered problem sets the solver trains on. Higher is better for both. Standard errors bootstrapped from 100 resamples. Best per scale in \best{green bold}.}
\label{tab:nli_diversity}
\small
\setlength{\tabcolsep}{5pt}
\begin{tabular}{ll cc cc}
\toprule
& & \multicolumn{2}{c}{\textbf{Semantic} $\uparrow$} & \multicolumn{2}{c}{\textbf{Logical} $\uparrow$} \\
\cmidrule(lr){3-4} \cmidrule(lr){5-6}
\textbf{Model} & \textbf{Setting} & Value & $\Delta$ & Value & $\Delta$ \\
\midrule
\multirow{2}{*}{Qwen3-4B}
& Baseline & $0.665 \pm {<}0.001$ & --- & $+0.241 \pm 0.013$ & --- \\
& \cellcolor{oursrow} VD & \cellcolor{oursrow}$\best{0.692} \pm {<}0.001$ & \cellcolor{oursrow}$+4.1\%$ & \cellcolor{oursrow}$\best{+0.405} \pm 0.014$ & \cellcolor{oursrow}$+68.1\%$ \\
\midrule
\multirow{2}{*}{Qwen3-8B}
& Baseline & $0.638 \pm {<}0.001$ & --- & $+0.315 \pm 0.013$ & --- \\
& \cellcolor{oursrow} VD & \cellcolor{oursrow}$\best{0.677} \pm {<}0.001$ & \cellcolor{oursrow}$+6.1\%$ & \cellcolor{oursrow}$\best{+0.376} \pm 0.016$ & \cellcolor{oursrow}$+19.4\%$ \\
\bottomrule
\end{tabular}
\end{table}

\subsection{Phase ablation embedding analysis}
\label{app:embedding_ablations}

\Cref{tab:question_diversity} tracks cumulative Vendi scores as questions are pooled across iterations, and \Cref{fig:embedding_ablations} shows per-iteration trends for all three embedding metrics. Gen-only dropout starts with high diversity but plateaus quickly (+4.6/+2.6 growth at 4B/8B), suggesting that, without a regularized solver, the proposer's distribution converges. Train-only starts lower but sustains steady growth (+9.6/+4.2), indicating that the regularized solver feeds back different reward signals that prevent proposer collapse. Both phases combined yield the highest cumulative diversity and the strongest growth (+8.7/+6.1), consistent with the solver accuracy results in \Cref{tab:vanilla_eval}.

\begin{table}[t]
\centering
\caption{Cumulative question diversity (Vendi score) over co-evolution iterations via \texttt{text-embedding-3-small} embeddings. \emph{Growth} is the gain from iteration 1 to iterations 1--5 pooled.}
\label{tab:question_diversity}
\vspace{0.3em}
\small
\setlength{\tabcolsep}{4pt}
\begin{tabular}{ll ccccc c}
\toprule
& & \multicolumn{5}{c}{\textbf{Cumulative Vendi Score} $\uparrow$} & \\
\cmidrule(lr){3-7}
\textbf{Model} & \textbf{Setting} & It.\ 1 & 1--2 & 1--3 & 1--4 & 1--5 & \textbf{Growth} \\
\midrule
\multirow{4}{*}{Qwen3-4B}
& Baseline           & 44.6 & 46.1 & 48.9 & 49.5 & 52.1 & +7.5 \\
& Train-only          & 46.1 & 51.0 & 52.0 & 53.7 & 55.7 & +9.6 \\
& Gen-only            & 59.7 & 62.9 & 63.9 & 63.9 & 64.3 & +4.6 \\
& \cellcolor{oursrow} Both & \cellcolor{oursrow}61.5 & \cellcolor{oursrow}63.9 & \cellcolor{oursrow}68.6 & \cellcolor{oursrow}69.7 & \cellcolor{oursrow}\best{70.2} & \cellcolor{oursrow}\best{+8.7} \\
\midrule
\multirow{4}{*}{Qwen3-8B}
& Baseline           & 40.0 & 41.5 & 42.1 & 43.2 & 42.2 & +2.2 \\
& Train-only          & 41.9 & 42.6 & 43.5 & 44.2 & 46.1 & +4.2 \\
& Gen-only            & 52.9 & 54.1 & 55.2 & 55.3 & 55.5 & +2.6 \\
& \cellcolor{oursrow} Both & \cellcolor{oursrow}51.1 & \cellcolor{oursrow}53.7 & \cellcolor{oursrow}56.0 & \cellcolor{oursrow}57.0 & \cellcolor{oursrow}\best{57.2} & \cellcolor{oursrow}\best{+6.1} \\
\bottomrule
\end{tabular}
\end{table}

\input{figures/embedding_ablations}

\subsection{Cross-scale co-evolution}
\label{app:cross_scale}

All experiments in \S\ref{sec:results} pair each model with itself (4B$\to$4B, 8B$\to$8B). A natural question is whether a stronger proposer generates a better curriculum for a weaker solver. We test this by pairing a Qwen3-8B proposer with a Qwen3-4B solver, with and without vocabulary dropout ($\alpha{=}0.85$, gen-only) on the 8B proposer.

\input{tables/cross_scale}

The cross-scale baseline (8B$\to$4B, no dropout) performs comparably to the symmetric baseline (4B$\to$4B), at 38.8 vs.\ 38.3 avg. A stronger proposer does not automatically produce a better curriculum, likely because the self-consistency reward calibrates difficulty to the solver's frontier regardless of proposer capacity.

Adding vocabulary dropout to the 8B proposer hurts rather than helps (37.4 vs.\ 38.8 avg), with a particularly large drop on AMC ($-8.4$) and AIME'25 ($-4.5$). This contrasts with the symmetric setting, where VD85 improves the 4B solver by +1.0 avg. This is consistent with the scale-dependent findings in \S\ref{sec:results}: vocabulary dropout addresses proposer stagnation during iterative co-adaptation, but when the proposer already has substantially more capacity than the solver, the difficulty calibration is misaligned and the extra diversity pressure makes the mismatch worse. This negative result supports the view that vocabulary dropout is a targeted intervention for co-evolutionary dynamics between capacity-matched models rather than a generic regularizer.

\subsection{Instruction-tuned models}
\label{app:instruct}

We use base (pre-trained) models throughout, consistent with R-Zero~\citep{huang2025rzero} and General-Reasoner~\citep{ma2025general}. We additionally run the pipeline on Qwen2.5-1.5B-Instruct (\Cref{tab:instruct}). No configuration consistently improves over the untrained base on either benchmark. At the tested scale, co-evolutionary training, with or without vocabulary dropout, does not benefit instruction-tuned models in our experiments.

\input{tables/instruct}

\subsection{Vocabulary utilization}
\label{app:vocab_utilization}

\Cref{tab:vocab_util} reports the number of unique tokens used in generated questions at iteration 5. Despite having fewer tokens available per batch, VD75 proposers use substantially more of the vocabulary than the baseline. This is consistent with the non-stationary mask forcing the proposer to explore a wider range of token compositions rather than concentrating on a fixed subset.

\begin{table}[H]
\centering
\caption{Unique tokens in iteration-5 generated questions.}
\label{tab:vocab_util}
\small
\setlength{\tabcolsep}{6pt}
\begin{tabular}{l cc cc}
\toprule
& \multicolumn{2}{c}{\textbf{Qwen3-4B}} & \multicolumn{2}{c}{\textbf{Qwen3-8B}} \\
\cmidrule(lr){2-3} \cmidrule(lr){4-5}
\textbf{Setting} & Unique tok. & \# Q & Unique tok. & \# Q \\
\midrule
Baseline & 2{,}717 & 6{,}771 & 1{,}881 & 7{,}044 \\
\rowcolor{oursrow} VD75 & \best{3{,}697} & 3{,}377 & \best{2{,}868} & 4{,}864 \\
\midrule
$\Delta$ & +36\% & & +52\% & \\
\bottomrule
\end{tabular}
\end{table}

\subsection{Curriculum-transfer training details}
\label{app:curriculum_transfer}

This expands the curriculum-transfer experiment of \S\ref{sec:curriculum_transfer}. We take the iteration-5 question sets from the baseline, KL-Cov, Clip-Cov, and VD ($\alpha{=}0.75$) 8B runs, verify every problem with an external model (GPT-5-mini, greedy decoding, single attempt), and keep only those whose independently-derived answer matches the curriculum's under \texttt{math\_verify} equivalence. External verification keeps ${\sim}20$--$31\%$ of each curriculum. We then GRPO-train Qwen3-4B-Base for 3 epochs on each verified subset using \texttt{prime-rl}~\citep{primeintellect2025prime-rl} (learning rate $1{\times}10^{-6}$, global batch size 32, group size 8, sequence length 2048, maximum completion length 1536, FSDP with DP=2 and a vLLM inference worker). Each checkpoint is evaluated pass@1 under the protocol of \Cref{tab:vanilla_eval}, with Minerva Math added.

\subsection{Annealed masking schedule}
\label{app:annealing}

\input{figures/annealing}

The scale-dependence of vocabulary dropout (\Cref{tab:vanilla_eval}) points toward adaptive masking schedules. We test a linear schedule increasing $\alpha$ from $0.75$ to $1.0$ over the 5 iterations (\Cref{fig:annealing}). The anneal peaks two iterations earlier than fixed VD75 (iteration 3 vs.\ 5), wins 3 of 5 iterations on math-average, and achieves the highest cumulative performance across all iterations. It also produces a more stable trajectory, with lower iteration-to-iteration volatility and an OlympiadBench range of 3.0 points vs.\ 6.4 for fixed (\Cref{fig:annealing}c). These results suggest applying diversity pressure early, when stagnation risk is highest, then relaxing it as the solver matures. We leave this direction to future work.

\subsection{Vocabulary dropout outside co-evolution}
\label{app:vanilla_grpo}

This section gives the setup and per-benchmark results behind the single-policy experiment in \S\ref{sec:generality}. We train Qwen3-4B-Base on GSM8K~\citep{cobbe2021gsm8k} with a binary correctness reward, with and without vocabulary dropout (a light mask, $\alpha{=}0.95$) applied to the policy during rollouts, and evaluate zero-shot on seven math benchmarks under the protocol of \Cref{tab:vanilla_eval}.

\begin{table}[H]
\centering
\caption{Vocabulary dropout applied to a single GRPO policy, with no co-evolution: Qwen3-4B-Base trained on GSM8K with a binary correctness reward, with ($\alpha{=}0.95$) and without ($\alpha{=}1.0$) vocabulary dropout, evaluated zero-shot (pass@1 \%). Best per benchmark in \best{green bold}.}
\label{tab:vanilla_grpo}
\small
\setlength{\tabcolsep}{4pt}
\begin{tabular}{l ccccccc}
\toprule
\textbf{Setting} & MATH500 & GSM8K & AMC & Olympiad & AIME'24 & AIME'25 & Minerva \\
\midrule
Baseline ($\alpha{=}1.0$) & \best{70.6} & 86.3 & \best{47.5} & 23.6 & 6.7 & \best{3.3} & 29.8 \\
\rowcolor{oursrow} VD ($\alpha{=}0.95$) & 63.0 & \best{87.9} & 45.0 & \best{24.1} & \best{10.0} & 0.0 & \best{34.9} \\
\bottomrule
\end{tabular}
\end{table}

Vocabulary dropout wins on four of the seven benchmarks (GSM8K, OlympiadBench, AIME 2024, Minerva), and the two arms are close on most tasks. The entropy trajectories for both arms are reported in \Cref{tab:vanilla_entropy}.

\section{Hyperparameters}
\label{app:hyperparameters}

\subsection{Evaluation benchmarks}
\label{app:benchmarks}

\input{tables/benchmarks}

\subsection{Training hyperparameters}
\label{app:training_hypers}

\begin{table}[H]
\centering
\caption{GRPO training hyperparameters for proposer and solver.}
\label{tab:training_hypers}
\scriptsize
\setlength{\tabcolsep}{4pt}
\begin{tabular}{lll}
\toprule
Parameter & Proposer & Solver \\
\midrule
Base model & \multicolumn{2}{c}{Qwen3-4B / Qwen3-8B} \\
Optimizer & \multicolumn{2}{c}{AdamW} \\
Learning rate & \multicolumn{2}{c}{$1 \times 10^{-6}$} \\
Weight decay & \multicolumn{2}{c}{$1 \times 10^{-2}$} \\
LR warmup & \multicolumn{2}{c}{0} \\
Max grad norm & \multicolumn{2}{c}{1.0} \\
KL penalty & \multicolumn{2}{c}{low-variance KL, $\beta = 10^{-2}$} \\
Gradient checkpointing & \multicolumn{2}{c}{enabled} \\
Precision & \multicolumn{2}{c}{bfloat16} \\
\midrule
Global batch size & 16 & 8 \\
Micro-batch (update) & 2 & 2 \\
Micro-batch (experience) & 4 & 4 \\
Rollout batch size & 16 & 8 \\
GRPO samples per prompt ($G$) & 4 & {--} \\
Self-consistency samples ($M$) & 10 & {--} \\
Temperature & 1.0 & 1.0 \\
Top-$p$ & 0.99 & 0.99 \\
Max response length & 2048 & 2048 \\
Max steps per phase & 6 & 20 \\
Epochs per phase & 10 & 10 \\
Co-evolution iterations & \multicolumn{2}{c}{5} \\
\midrule
Difficulty window $[\tau_{\min}, \tau_{\max}]$ & \multicolumn{2}{c}{$[0.3, 0.7]$} \\
Vocab dropout $\alpha$ & 0.75 & {--} \\
BLEU rep.\ penalty $\tau_{\text{BLEU}}$ & 0.5 & {--} \\
\bottomrule
\end{tabular}
\end{table}

\subsection{Vendi score computation}
\label{app:vendi}

We compute the Vendi score~\citep{friedman2023vendi} for each (experiment, iteration) pair by embedding the proposer's generated questions and computing the eigenspectrum of the cosine similarity kernel. Given $n$ questions with L2-normalized embeddings $E \in \mathbb{R}^{n \times d}$, the similarity matrix is $K = E E^\top$. We compute $\text{VS} = \exp(-\sum_i \hat{\lambda}_i \log \hat{\lambda}_i)$ where $\hat{\lambda}_i = \lambda_i / \sum_j \lambda_j$ are the normalized eigenvalues of $K$. This yields the effective number of distinct question types.

\begin{table}[H]
\centering
\caption{Vendi score hyperparameters.}
\label{tab:vendi_hypers}
\scriptsize
\setlength{\tabcolsep}{4pt}
\begin{tabular}{ll}
\toprule
Parameter & Value \\
\midrule
Embedding model & \texttt{text-embedding-3-small} \\
Embedding dimension & 1536 \\
Normalization & L2 (cosine kernel) \\
Max samples (eigendecomp.) & 5{,}000 \\
Subsampling seed & 42 \\
Eigenvalue threshold & $10^{-12}$ \\
\bottomrule
\end{tabular}
\end{table}

\subsection{Epiplexity computation}
\label{app:epiplexity}

Our epiplexity computation follows \citet{finzi2026entropy}, using the prequential MDL procedure that \citet{liu2026selfplay} applied to self-play. For each (experiment, iteration) pair we fine-tune a fresh LoRA observer on the proposer's generated questions and ask how much structure it can extract, defined as the gap between the online loss on each example before training and the converged training loss, in bits per token. A curriculum with more learnable structure yields higher epiplexity.

The observer starts from the same base model used in the co-evolutionary loop, with LoRA adapters on all attention and MLP projections. Training runs an online-then-converge protocol, where epoch 1 records the loss on each batch before updating (the prequential code length) and training then continues for up to 20 epochs with early stopping on a held-out validation set.

\begin{table}[H]
\centering
\caption{Epiplexity computation hyperparameters.}
\label{tab:epiplexity_hypers}
\scriptsize
\setlength{\tabcolsep}{4pt}
\begin{tabular}{ll}
\toprule
Parameter & Value \\
\midrule
Observer model & (same as base) \\
LoRA rank ($r$) & 16 \\
LoRA alpha & 32 \\
LoRA dropout & 0.05 \\
Target modules & q/k/v/o\_proj, gate/up/down\_proj \\
Max sequence length & 512 \\
Batch size & 64 \\
Learning rate & $10^{-4}$ \\
Optimizer & AdamW (weight decay 0.01) \\
Max epochs & 20 \\
Early stopping patience & 5 epochs \\
Validation split & 10\% \\
Min question length & 20 characters \\
Precision & bfloat16 \\
\bottomrule
\end{tabular}
\end{table}

Epiplexity per token is computed as $(\mathcal{L}_{\text{online}} - \mathcal{L}_{\text{train}}) / \ln 2$, where $\mathcal{L}_{\text{online}}$ is the total cross-entropy loss from the prequential (epoch 1) pass and $\mathcal{L}_{\text{train}}$ is the total loss at the best MDL epoch. The MDL criterion selects the epoch that minimizes $\text{epiplexity}/N_{\text{train}} + \mathcal{L}_{\text{val}}/(\ln 2 \cdot N_{\text{val}})$.

\section{Prompt templates}
\label{app:prompts}

\subsection{Original R-Zero questioner prompt}
\label{app:rzero_prompt}

The following is the out-of-the-box R-Zero questioner prompt, which targets various fields of mathematics at competition level. All results in \S\ref{sec:results} use this prompt.

\begin{tcolorbox}[promptbox, title={\color{white}R-Zero Questioner (System Message)}]
You are an expert competition-math problem setter.\par
FIRST, in your private scratch-pad, think step-by-step to design a brand-new, non-trivial problem. The problem could come from any field of mathematics, including but not limited to algebra, geometry, number theory, combinatorics, prealgebra, probability, statistics, and calculus. Aim for a difficulty such that fewer than 30\% of advanced high-school students could solve it. Avoid re-using textbook clich\'{e}s or famous contest problems.\par
\medskip
THEN, without revealing any of your private thoughts, output \textbf{exactly} the following two blocks:\par
\medskip
<question>\par
\{The full problem statement on one or more lines\}\par
</question>\par
\medskip
\textbackslash boxed\{final\_answer\}\par
\medskip
Do NOT output anything else: no explanations, no extra markup.
\end{tcolorbox}

\begin{tcolorbox}[promptbox, title={\color{white}R-Zero Questioner (User Message)}]
Generate one new, challenging reasoning question now. Remember to format the output exactly as instructed.
\end{tcolorbox}

\subsection{Evaluation prompt}
\label{app:eval_prompt}

All benchmarks are evaluated zero-shot using the following user message, with no system prompt:

\begin{tcolorbox}[promptbox, title={\color{white}Evaluation (User Message)}]
\{question\}\par
Let's think step by step and provide your final answer inside \textbackslash boxed\{\} notation.
\end{tcolorbox}

The question text is passed verbatim from each dataset's problem field. The prompt is wrapped in the model's native chat template (e.g., \texttt{<|im\_start|>user} \ldots\ \texttt{<|im\_end|>} for Qwen3 models). No system message is set.

\end{document}

%% file: tables/vanilla_eval.tex
\begin{table*}[t]
\caption{Solver pass@1 accuracy (\%) after 5 co-evolution iterations. Vocabulary dropout (VD) is applied to the \emph{proposer} only, and the solver trains on the resulting curriculum. $\alpha$ controls the fraction of the vocabulary retained. Phase ablations decompose $\alpha{=}0.75$ into train-only and gen-only. Best per model in \best{green bold}.}
\label{tab:vanilla_eval}
\vspace{0.3em}
\centering
\footnotesize
\setlength{\tabcolsep}{5pt}
\begin{tabular}{l cccccc c}
\toprule
& \multicolumn{6}{c}{\textbf{Pass@1 (\%)}} & \\
\cmidrule(lr){2-7}
\textbf{Setting} & MATH500 & GSM8K & AMC & Olympiad & AIME'24 & AIME'25 & \textbf{Avg.} \\
\midrule
\multicolumn{8}{c}{\textbf{Qwen3-4B}} \\
\midrule
\textcolor{gray}{Base (no training)} & \textcolor{gray}{54.1\tss{2.2}} & \textcolor{gray}{63.5\tss{5.2}} & \textcolor{gray}{29.2\tss{1.8}} & \textcolor{gray}{20.6\tss{1.3}} & \textcolor{gray}{12.2\tss{0.9}} & \textcolor{gray}{1.1\tss{0.9}} & \textcolor{gray}{30.1} \\
KL-Cov & 60.8\tss{1.9} & 82.6\tss{2.0} & 42.0\tss{3.8} & 23.0\tss{0.8} & 9.3\tss{2.6} & 4.0\tss{1.1} & 37.0 \\
Clip-Cov & 65.8\tss{1.1} & 85.4\tss{1.0} & 39.5\tss{3.4} & 24.5\tss{0.2} & \best{12.7}\tss{1.7} & 4.0\tss{1.7} & 38.7 \\
Baseline ($\alpha{=}1.0$) & 64.2\tss{0.8} & 85.2\tss{0.7} & 41.7\tss{1.4} & 23.2\tss{0.3} & 8.9\tss{0.9} & 6.7\tss{1.6} & 38.3 \\
\rowcolor{oursrow} $\alpha{=}0.85$ & 65.7\tss{0.9} & 81.8\tss{0.4} & \best{50.0}\tss{2.4} & 24.7\tss{0.9} & 8.9\tss{1.8} & 4.4\tss{0.9} & \best{39.3} \\
\rowcolor{oursrow} $\alpha{=}0.75$ & \best{66.0}\tss{0.7} & 82.4\tss{1.6} & 35.8\tss{5.9} & \best{24.8}\tss{0.1} & 5.6\tss{1.8} & 4.4\tss{1.8} & 36.5 \\
\cmidrule(lr){1-8}
\rowcolor{oursrow} \quad train-only & 64.5\tss{0.4} & 85.4\tss{1.1} & 31.7\tss{1.8} & 23.6\tss{0.2} & 11.1\tss{3.3} & \best{7.8}\tss{1.8} & 37.4 \\
\rowcolor{oursrow} \quad gen-only & 62.9\tss{1.3} & \best{85.9}\tss{0.7} & 42.5\tss{1.2} & 24.1\tss{0.8} & 7.8\tss{0.9} & 3.3\tss{1.6} & 37.8 \\
\midrule
\multicolumn{8}{c}{\textbf{Qwen3-8B}} \\
\midrule
\textcolor{gray}{Base (no training)} & \textcolor{gray}{70.7\tss{0.1}} & \textcolor{gray}{82.2\tss{0.7}} & \textcolor{gray}{45.8\tss{3.6}} & \textcolor{gray}{18.1\tss{1.3}} & \textcolor{gray}{8.9\tss{2.4}} & \textcolor{gray}{7.8\tss{0.9}} & \textcolor{gray}{38.9} \\
KL-Cov & \best{71.1}\tss{1.7} & 88.8\tss{0.5} & 45.0\tss{2.6} & 26.9\tss{0.8} & 8.7\tss{2.0} & 9.3\tss{1.1} & 41.6 \\
Clip-Cov & 70.0\tss{1.3} & 86.5\tss{2.6} & 42.5\tss{2.0} & 28.0\tss{0.5} & 12.7\tss{1.7} & 9.3\tss{1.7} & 41.5 \\
Baseline ($\alpha{=}1.0$) & 69.9\tss{0.4} & 86.9\tss{2.5} & 40.8\tss{7.7} & 23.1\tss{1.8} & 12.2\tss{0.9} & 3.3\tss{1.6} & 39.4 \\
\rowcolor{oursrow} $\alpha{=}0.85$ & 69.7\tss{2.3} & 89.5\tss{0.8} & 45.8\tss{1.8} & 26.8\tss{0.5} & 6.7\tss{0.0} & 7.8\tss{3.3} & 41.1 \\
\rowcolor{oursrow} $\alpha{=}0.75$ & 68.1\tss{2.3} & 88.6\tss{1.4} & \best{51.7}\tss{3.6} & 27.7\tss{0.3} & \best{15.6}\tss{2.4} & \best{11.1}\tss{2.4} & \best{43.8} \\
\cmidrule(lr){1-8}
\rowcolor{oursrow} \quad train-only & \best{71.1}\tss{1.5} & \best{89.7}\tss{0.6} & 44.2\tss{3.6} & 27.0\tss{1.1} & 13.3\tss{0.0} & 5.6\tss{2.4} & 41.8 \\
\rowcolor{oursrow} \quad gen-only & 71.0\tss{2.6} & 87.0\tss{1.2} & 49.2\tss{4.1} & \best{29.2}\tss{0.7} & 11.1\tss{0.9} & 5.6\tss{2.4} & 42.2 \\
\bottomrule
\end{tabular}
\end{table*}

%% file: figures/question_profile.tex
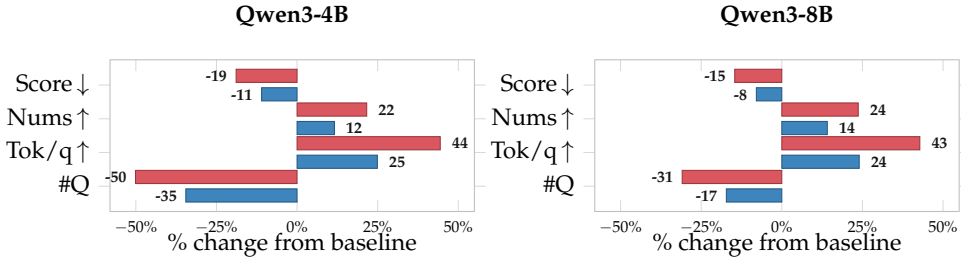
\begin{figure}[H]
\centering

\pgfplotsset{
  qprofile/.style={
    xbar,
    width=0.46\columnwidth,
    height=3.6cm,
    bar width=5pt,
    xmin=-58, xmax=55,
    xlabel={\small\% change from baseline},
    xlabel style={font=\scriptsize, at={(axis description cs:0.5,-0.20)}},
    ytick=data,
    yticklabels={\#Q, Tok/q\,$\uparrow$, Nums\,$\uparrow$, Score\,$\downarrow$},
    yticklabel style={font=\footnotesize},
    xticklabel style={font=\scriptsize},
    xtick={-50,-25,0,25,50},
    xticklabel={\pgfmathprintnumber{\tick}\%},
    enlarge y limits=0.25,
    every axis plot/.append style={fill opacity=0.92},
    axis line style={gray!50},
    tick style={gray!30},
    xmajorgrids=false,
    ymajorgrids=false,
    extra x ticks={0},
    extra x tick style={grid style={black!30, semithick}, major tick length=0pt},
    extra x tick labels={},
    clip=false,
    nodes near coords,
    nodes near coords style={
      font=\scriptsize\bfseries,
      anchor=west,
      xshift=1pt,
      /pgf/number format/fixed,
      /pgf/number format/precision=0,
      /pgf/number format/assume math mode=true,
    },
    every node near coord/.append style={
      /utils/exec={%
        \pgfmathfloatifflags{\pgfplotspointmeta}{2}{%
          \pgfkeysalso{anchor=east, xshift=-1pt}%
        }{}%
      },
    },
  }
}

\begin{tikzpicture}
\begin{groupplot}[
  group style={
    group size=2 by 1,
    horizontal sep=1.6cm,
  },
  qprofile,
]

\nextgroupplot[title={\small\textbf{Qwen3-4B}}, title style={at={(0.5,1.05)}}]
\addplot[fill=vd85col, draw=vd85col!70!black, line width=0.3pt] coordinates
  {(-34.5,0) (24.9,1) (11.6,2) (-11.0,3)};
\addplot[fill=vd75col, draw=vd75col!70!black, line width=0.3pt] coordinates
  {(-50.1,0) (44.4,1) (21.6,2) (-18.9,3)};

\nextgroupplot[title={\small\textbf{Qwen3-8B}}, title style={at={(0.5,1.05)}}]
\addplot[fill=vd85col, draw=vd85col!70!black, line width=0.3pt] coordinates
  {(-17.2,0) (24.0,1) (14.1,2) (-7.9,3)};
\addplot[fill=vd75col, draw=vd75col!70!black, line width=0.3pt] coordinates
  {(-30.9,0) (42.8,1) (23.7,2) (-14.6,3)};

\end{groupplot}
\end{tikzpicture}
\caption{Question profile at iteration 5 (\% change from baseline). \legendsquare{vd85col}\,VD85 \legendsquare{vd75col}\,VD75.}
\label{fig:question_complexity}
\end{figure}

%% file: figures/diversity_combined.tex
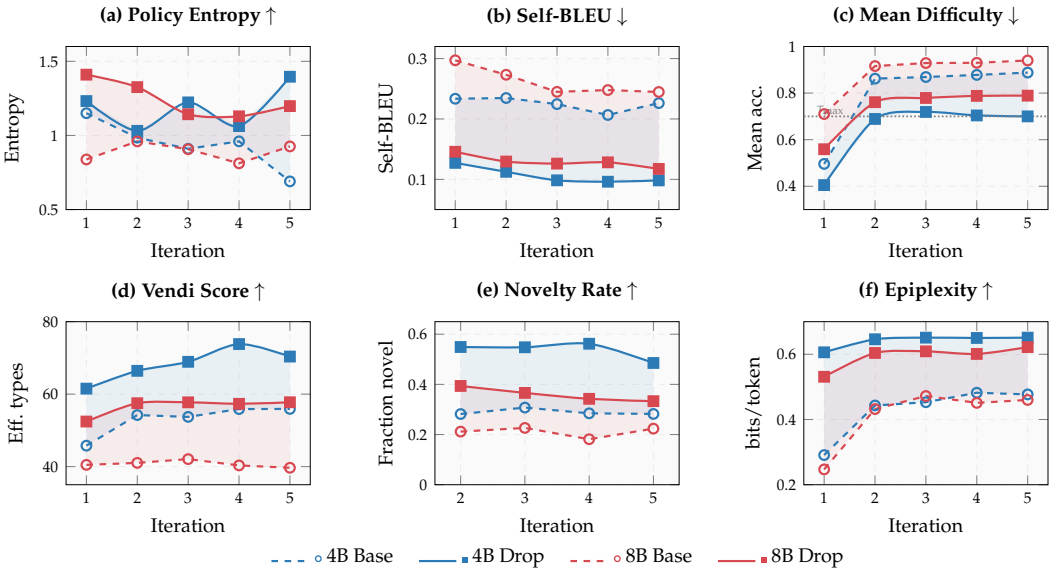
\begin{figure*}[t]
\centering
\resizebox{\textwidth}{!}{%
\begin{tikzpicture}

\definecolor{color4b}{RGB}{76,114,176}
\definecolor{color8b}{RGB}{221,132,82}
\definecolor{fill4b}{RGB}{76,114,176}
\definecolor{fill8b}{RGB}{221,132,82}

\begin{groupplot}[
    group style={
        group size=3 by 2,
        horizontal sep=1.5cm,
        vertical sep=1.8cm,
        group name=divplots
    },
    width=5.5cm,
    height=3.8cm,
    xmin=0.6, xmax=5.4,
    xtick={1,2,3,4,5},
    xlabel style={font=\large},
    ylabel style={font=\large},
    tick label style={font=\normalsize},
    title style={font=\large\bfseries},
    grid=major,
    grid style={gray!15, dashed},
    every axis plot/.append style={line width=1.0pt, mark size=2.2pt},
    axis background/.style={fill=gray!4},
]


\nextgroupplot[
    title={(a) Policy Entropy $\uparrow$},
    ylabel={Entropy},
    ymin=0.5, ymax=1.6,
    xlabel={Iteration},
]
\addplot[color4b, dashed, mark=o, mark options={solid}] coordinates {
    (1, 1.150) (2, 0.985) (3, 0.910) (4, 0.960) (5, 0.690)
};
\addplot[color4b, mark=square*, smooth] coordinates {
    (1, 1.232) (2, 1.031) (3, 1.223) (4, 1.063) (5, 1.396)
};
\addplot[name path=4b_base_a, draw=none, forget plot] coordinates {(1, 1.150) (2, 0.985) (3, 0.910) (4, 0.960) (5, 0.690)};
\addplot[name path=4b_drop_a, draw=none, forget plot] coordinates {(1, 1.232) (2, 1.031) (3, 1.223) (4, 1.063) (5, 1.396)};
\addplot[fill=fill4b, fill opacity=0.08, forget plot] fill between[of=4b_base_a and 4b_drop_a];
\addplot[color8b, dashed, mark=o, mark options={solid}] coordinates {
    (1, 0.838) (2, 0.960) (3, 0.907) (4, 0.812) (5, 0.926)
};
\addplot[color8b, mark=square*, smooth] coordinates {
    (1, 1.410) (2, 1.326) (3, 1.143) (4, 1.129) (5, 1.198)
};
\addplot[name path=8b_base_a, draw=none, forget plot] coordinates {(1, 0.838) (2, 0.960) (3, 0.907) (4, 0.812) (5, 0.926)};
\addplot[name path=8b_drop_a, draw=none, forget plot] coordinates {(1, 1.410) (2, 1.326) (3, 1.143) (4, 1.129) (5, 1.198)};
\addplot[fill=fill8b, fill opacity=0.08, forget plot] fill between[of=8b_base_a and 8b_drop_a];

\nextgroupplot[
    title={(b) Self-BLEU $\downarrow$},
    ylabel={Self-BLEU},
    ymin=0.05, ymax=0.32,
    xlabel={Iteration},
]
\addplot[color4b, dashed, mark=o, mark options={solid}] coordinates {
    (1, 0.2335) (2, 0.2347) (3, 0.2246) (4, 0.2066) (5, 0.2263)
};
\addplot[color4b, mark=square*, smooth] coordinates {
    (1, 0.1273) (2, 0.1125) (3, 0.0984) (4, 0.0961) (5, 0.0983)
};
\addplot[name path=4b_base_b, draw=none, forget plot] coordinates {(1, 0.2335) (2, 0.2347) (3, 0.2246) (4, 0.2066) (5, 0.2263)};
\addplot[name path=4b_drop_b, draw=none, forget plot] coordinates {(1, 0.1273) (2, 0.1125) (3, 0.0984) (4, 0.0961) (5, 0.0983)};
\addplot[fill=fill4b, fill opacity=0.08, forget plot] fill between[of=4b_base_b and 4b_drop_b];
\addplot[color8b, dashed, mark=o, mark options={solid}] coordinates {
    (1, 0.2971) (2, 0.2732) (3, 0.2449) (4, 0.2479) (5, 0.2447)
};
\addplot[color8b, mark=square*, smooth] coordinates {
    (1, 0.1457) (2, 0.1295) (3, 0.1261) (4, 0.1282) (5, 0.1173)
};
\addplot[name path=8b_base_b, draw=none, forget plot] coordinates {(1, 0.2971) (2, 0.2732) (3, 0.2449) (4, 0.2479) (5, 0.2447)};
\addplot[name path=8b_drop_b, draw=none, forget plot] coordinates {(1, 0.1457) (2, 0.1295) (3, 0.1261) (4, 0.1282) (5, 0.1173)};
\addplot[fill=fill8b, fill opacity=0.08, forget plot] fill between[of=8b_base_b and 8b_drop_b];

\nextgroupplot[
    title={(c) Mean Difficulty $\downarrow$},
    ylabel={Mean acc.},
    ymin=0.3, ymax=1.0,
    xlabel={Iteration},
]
\addplot[color4b, dashed, mark=o, mark options={solid}] coordinates {
    (1, 0.496) (2, 0.862) (3, 0.869) (4, 0.878) (5, 0.888)
};
\addplot[color4b, mark=square*, smooth] coordinates {
    (1, 0.405) (2, 0.689) (3, 0.719) (4, 0.704) (5, 0.700)
};
\addplot[name path=4b_base_c, draw=none, forget plot] coordinates {(1, 0.496) (2, 0.862) (3, 0.869) (4, 0.878) (5, 0.888)};
\addplot[name path=4b_drop_c, draw=none, forget plot] coordinates {(1, 0.405) (2, 0.689) (3, 0.719) (4, 0.704) (5, 0.700)};
\addplot[fill=fill4b, fill opacity=0.08, forget plot] fill between[of=4b_base_c and 4b_drop_c];
\addplot[color8b, dashed, mark=o, mark options={solid}] coordinates {
    (1, 0.710) (2, 0.916) (3, 0.929) (4, 0.930) (5, 0.940)
};
\addplot[color8b, mark=square*, smooth] coordinates {
    (1, 0.559) (2, 0.761) (3, 0.779) (4, 0.788) (5, 0.789)
};
\addplot[name path=8b_base_c, draw=none, forget plot] coordinates {(1, 0.710) (2, 0.916) (3, 0.929) (4, 0.930) (5, 0.940)};
\addplot[name path=8b_drop_c, draw=none, forget plot] coordinates {(1, 0.559) (2, 0.761) (3, 0.779) (4, 0.788) (5, 0.789)};
\addplot[fill=fill8b, fill opacity=0.08, forget plot] fill between[of=8b_base_c and 8b_drop_c];
\draw[gray, densely dotted, line width=0.8pt] (axis cs:0.6,0.7) -- (axis cs:5.4,0.7);
\node[anchor=east, font=\footnotesize, gray] at (axis cs:5.3,0.745) {$\tau_{\max}$};


\nextgroupplot[
    title={(d) Vendi Score $\uparrow$},
    ylabel={Eff.\ types},
    ymin=35, ymax=80,
    xlabel={Iteration},
]
\addplot[color4b, dashed, mark=o, mark options={solid}] coordinates {
    (1, 45.78) (2, 54.23) (3, 53.72) (4, 55.88) (5, 55.92)
};
\addplot[color4b, mark=square*, smooth] coordinates {
    (1, 61.52) (2, 66.43) (3, 68.92) (4, 73.82) (5, 70.37)
};
\addplot[name path=4b_base_d, draw=none, forget plot] coordinates {(1, 45.78) (2, 54.23) (3, 53.72) (4, 55.88) (5, 55.92)};
\addplot[name path=4b_drop_d, draw=none, forget plot] coordinates {(1, 61.52) (2, 66.43) (3, 68.92) (4, 73.82) (5, 70.37)};
\addplot[fill=fill4b, fill opacity=0.08, forget plot] fill between[of=4b_base_d and 4b_drop_d];
\addplot[color8b, dashed, mark=o, mark options={solid}] coordinates {
    (1, 40.48) (2, 41.03) (3, 42.07) (4, 40.33) (5, 39.68)
};
\addplot[color8b, mark=square*, smooth] coordinates {
    (1, 52.44) (2, 57.50) (3, 57.74) (4, 57.33) (5, 57.76)
};
\addplot[name path=8b_base_d, draw=none, forget plot] coordinates {(1, 40.48) (2, 41.03) (3, 42.07) (4, 40.33) (5, 39.68)};
\addplot[name path=8b_drop_d, draw=none, forget plot] coordinates {(1, 52.44) (2, 57.50) (3, 57.74) (4, 57.33) (5, 57.76)};
\addplot[fill=fill8b, fill opacity=0.08, forget plot] fill between[of=8b_base_d and 8b_drop_d];

\nextgroupplot[
    title={(e) Novelty Rate $\uparrow$},
    ylabel={Fraction novel},
    ymin=0.0, ymax=0.65,
    xmin=1.6, xmax=5.4,
    xtick={2,3,4,5},
    xlabel={Iteration},
]
\addplot[color4b, dashed, mark=o, mark options={solid}] coordinates {
    (2, 0.2818) (3, 0.3075) (4, 0.2846) (5, 0.2822)
};
\addplot[color4b, mark=square*, smooth] coordinates {
    (2, 0.549) (3, 0.5479) (4, 0.562) (5, 0.4853)
};
\addplot[name path=4b_base_e, draw=none, forget plot] coordinates {(2, 0.2818) (3, 0.3075) (4, 0.2846) (5, 0.2822)};
\addplot[name path=4b_drop_e, draw=none, forget plot] coordinates {(2, 0.549) (3, 0.5479) (4, 0.562) (5, 0.4853)};
\addplot[fill=fill4b, fill opacity=0.08, forget plot] fill between[of=4b_base_e and 4b_drop_e];
\addplot[color8b, dashed, mark=o, mark options={solid}] coordinates {
    (2, 0.2117) (3, 0.2264) (4, 0.1814) (5, 0.2235)
};
\addplot[color8b, mark=square*, smooth] coordinates {
    (2, 0.3937) (3, 0.366) (4, 0.3424) (5, 0.3331)
};
\addplot[name path=8b_base_e, draw=none, forget plot] coordinates {(2, 0.2117) (3, 0.2264) (4, 0.1814) (5, 0.2235)};
\addplot[name path=8b_drop_e, draw=none, forget plot] coordinates {(2, 0.3937) (3, 0.366) (4, 0.3424) (5, 0.3331)};
\addplot[fill=fill8b, fill opacity=0.08, forget plot] fill between[of=8b_base_e and 8b_drop_e];

\nextgroupplot[
    title={(f) Epiplexity $\uparrow$},
    ylabel={bits/token},
    ymin=0.2, ymax=0.7,
    xlabel={Iteration},
]
\addplot[color4b, dashed, mark=o, mark options={solid}] coordinates {
    (1, 0.291) (2, 0.443) (3, 0.453) (4, 0.482) (5, 0.477)
};
\addplot[color4b, mark=square*, smooth] coordinates {
    (1, 0.606) (2, 0.646) (3, 0.651) (4, 0.650) (5, 0.651)
};
\addplot[name path=4b_base_f, draw=none, forget plot] coordinates {(1, 0.291) (2, 0.443) (3, 0.453) (4, 0.482) (5, 0.477)};
\addplot[name path=4b_drop_f, draw=none, forget plot] coordinates {(1, 0.606) (2, 0.646) (3, 0.651) (4, 0.650) (5, 0.651)};
\addplot[fill=fill4b, fill opacity=0.08, forget plot] fill between[of=4b_base_f and 4b_drop_f];
\addplot[color8b, dashed, mark=o, mark options={solid}] coordinates {
    (1, 0.247) (2, 0.432) (3, 0.472) (4, 0.451) (5, 0.460)
};
\addplot[color8b, mark=square*, smooth] coordinates {
    (1, 0.531) (2, 0.604) (3, 0.609) (4, 0.601) (5, 0.622)
};
\addplot[name path=8b_base_f, draw=none, forget plot] coordinates {(1, 0.247) (2, 0.432) (3, 0.472) (4, 0.451) (5, 0.460)};
\addplot[name path=8b_drop_f, draw=none, forget plot] coordinates {(1, 0.531) (2, 0.604) (3, 0.609) (4, 0.601) (5, 0.622)};
\addplot[fill=fill8b, fill opacity=0.08, forget plot] fill between[of=8b_base_f and 8b_drop_f];

\end{groupplot}

\path (divplots c1r2.south west) -- (divplots c3r2.south east) coordinate[midway, yshift=-0.9cm] (legendpos);
\node[anchor=north, font=\large] at (legendpos) {
    \tikz\draw[color4b, dashed, line width=1pt] (0,0) -- (0.6,0); \tikz\node[color4b,circle,draw,inner sep=0.9pt,line width=0.6pt] {}; 4B Base \quad
    \tikz\draw[color4b, line width=1pt] (0,0) -- (0.6,0); \tikz\node[color4b,fill=color4b,rectangle,inner sep=1.2pt] {}; 4B VD75 \quad
    \tikz\draw[color8b, dashed, line width=1pt] (0,0) -- (0.6,0); \tikz\node[color8b,circle,draw,inner sep=0.9pt,line width=0.6pt] {}; 8B Base \quad
    \tikz\draw[color8b, line width=1pt] (0,0) -- (0.6,0); \tikz\node[color8b,fill=color8b,rectangle,inner sep=1.2pt] {}; 8B VD75
};

\end{tikzpicture}%
}
\caption{Diversity and curriculum quality over co-evolution iterations ($\alpha{=}0.75$, both phases). Top: collapse signals. Bottom: diversity metrics. Dashed = baseline, solid = dropout. Semantic metrics (b, d, e) use \texttt{text-embedding-3-small}~\citep{openai2024embeddings}.}
\label{fig:diversity_combined}
\end{figure*}

%% file: figures/embedding_diversity.tex
\begin{figure}[H]
\centering
\begin{tikzpicture}

\definecolor{color4b}{RGB}{76,114,176}
\definecolor{color8b}{RGB}{221,132,82}

\pgfplotsset{
  embdiv/.style={
    width=0.75\columnwidth,
    height=5cm,
    xlabel={Iteration},
    xtick={1,2,3,4,5},
    xticklabel style={font=\small},
    yticklabel style={font=\small},
    xlabel style={font=\small},
    ylabel style={font=\small},
    ylabel={Effective \# distinct questions},
    ymin=35, ymax=75,
    xmin=0.8, xmax=5.2,
    grid=major,
    grid style={gray!20, very thin},
    mark size=2pt,
    line width=0.9pt,
    clip=false,
    legend style={
      font=\scriptsize,
      draw=none,
      fill=none,
      column sep=3pt,
      at={(0.5,-0.30)},
      anchor=north,
      legend columns=4,
    },
  }
}

\begin{axis}[embdiv]
\addplot[color4b, dashed, mark=o, mark options={solid}] coordinates {(1,44.6) (2,46.1) (3,48.9) (4,49.5) (5,52.1)};
\addplot[color8b, dashed, mark=o, mark options={solid}] coordinates {(1,40.0) (2,41.5) (3,42.1) (4,43.2) (5,42.2)};
\addplot[color4b, mark=square*, mark options={solid}] coordinates {(1,61.5) (2,63.9) (3,68.6) (4,69.7) (5,70.2)};
\addplot[color8b, mark=square*, mark options={solid}] coordinates {(1,51.1) (2,53.7) (3,56.0) (4,57.0) (5,57.2)};
\addlegendentry{4B Base}
\addlegendentry{8B Base}
\addlegendentry{4B VD75}
\addlegendentry{8B VD75}
\end{axis}

\end{tikzpicture}
\caption{Cumulative Vendi score (questions pooled across iterations 1--$N$, subsampled to 2{,}000). VD75 maintains higher diversity at both scales.}
\label{fig:embedding_diversity}
\end{figure}

%% file: figures/embedding_ablations.tex
\begin{figure*}[t]
\centering
\resizebox{\textwidth}{!}{%
\begin{tikzpicture}

\definecolor{cbase}{RGB}{120,120,120}
\definecolor{ctrain}{RGB}{76,114,176}
\definecolor{cgen}{RGB}{31,138,112}
\definecolor{cboth}{RGB}{221,132,82}

\begin{groupplot}[
    group style={
        group size=3 by 2,
        horizontal sep=1.5cm,
        vertical sep=1.6cm,
        group name=ablplots
    },
    width=5.2cm,
    height=4.0cm,
    xmin=0.6, xmax=5.4,
    xtick={1,2,3,4,5},
    xlabel={Iteration},
    xlabel style={font=\large},
    ylabel style={font=\large},
    tick label style={font=\normalsize},
    title style={font=\large\bfseries},
    grid=major,
    grid style={gray!15, dashed},
    every axis plot/.append style={line width=0.9pt, mark size=2pt},
    axis background/.style={fill=gray!4},
]


\nextgroupplot[
    title={(a) 4B Vendi Score $\uparrow$},
    ylabel={Eff.\ types},
    ymin=40, ymax=78,
]
\addplot[cbase, dashed, mark=o, mark options={solid}] coordinates {(1,45.78)(2,54.23)(3,53.72)(4,55.88)(5,55.92)};
\addplot[ctrain, mark=triangle*, mark options={solid}] coordinates {(1,48.86)(2,60.29)(3,63.38)(4,62.13)(5,63.99)};
\addplot[cgen, mark=diamond*, mark options={solid}] coordinates {(1,59.97)(2,63.25)(3,64.32)(4,62.59)(5,60.84)};
\addplot[cboth, mark=square*, mark options={solid}] coordinates {(1,61.52)(2,66.43)(3,68.92)(4,73.82)(5,70.37)};

\nextgroupplot[
    title={(b) 4B Self-BLEU $\downarrow$},
    ylabel={Self-BLEU},
    ymin=0.06, ymax=0.28,
]
\addplot[cbase, dashed, mark=o, mark options={solid}] coordinates {(1,0.2335)(2,0.2347)(3,0.2246)(4,0.2066)(5,0.2263)};
\addplot[ctrain, mark=triangle*, mark options={solid}] coordinates {(1,0.2458)(2,0.19)(3,0.1866)(4,0.2002)(5,0.1924)};
\addplot[cgen, mark=diamond*, mark options={solid}] coordinates {(1,0.1488)(2,0.1188)(3,0.1124)(4,0.1265)(5,0.1156)};
\addplot[cboth, mark=square*, mark options={solid}] coordinates {(1,0.1273)(2,0.1125)(3,0.0984)(4,0.0961)(5,0.0983)};

\nextgroupplot[
    title={(c) 4B Novelty Rate $\uparrow$},
    ylabel={Fraction novel},
    ymin=0.1, ymax=0.62,
    xmin=1.6, xmax=5.4,
    xtick={2,3,4,5},
]
\addplot[cbase, dashed, mark=o, mark options={solid}] coordinates {(2,0.2818)(3,0.3075)(4,0.2846)(5,0.2822)};
\addplot[ctrain, mark=triangle*, mark options={solid}] coordinates {(2,0.365)(3,0.3551)(4,0.3569)(5,0.3607)};
\addplot[cgen, mark=diamond*, mark options={solid}] coordinates {(2,0.5256)(3,0.4965)(4,0.454)(5,0.3813)};
\addplot[cboth, mark=square*, mark options={solid}] coordinates {(2,0.549)(3,0.5479)(4,0.562)(5,0.4853)};


\nextgroupplot[
    title={(d) 8B Vendi Score $\uparrow$},
    ylabel={Eff.\ types},
    ymin=35, ymax=60,
]
\addplot[cbase, dashed, mark=o, mark options={solid}] coordinates {(1,40.48)(2,41.03)(3,42.07)(4,40.33)(5,39.68)};
\addplot[ctrain, mark=triangle*, mark options={solid}] coordinates {(1,42.5)(2,42.65)(3,39.96)(4,42.29)(5,43.58)};
\addplot[cgen, mark=diamond*, mark options={solid}] coordinates {(1,53.44)(2,55.44)(3,54.44)(4,56.13)(5,54.2)};
\addplot[cboth, mark=square*, mark options={solid}] coordinates {(1,52.44)(2,57.5)(3,57.74)(4,57.33)(5,57.76)};

\nextgroupplot[
    title={(e) 8B Self-BLEU $\downarrow$},
    ylabel={Self-BLEU},
    ymin=0.06, ymax=0.32,
]
\addplot[cbase, dashed, mark=o, mark options={solid}] coordinates {(1,0.2971)(2,0.2732)(3,0.2449)(4,0.2479)(5,0.2447)};
\addplot[ctrain, mark=triangle*, mark options={solid}] coordinates {(1,0.2775)(2,0.2741)(3,0.2709)(4,0.2547)(5,0.2263)};
\addplot[cgen, mark=diamond*, mark options={solid}] coordinates {(1,0.1437)(2,0.1192)(3,0.1242)(4,0.116)(5,0.1253)};
\addplot[cboth, mark=square*, mark options={solid}] coordinates {(1,0.1457)(2,0.1295)(3,0.1261)(4,0.1282)(5,0.1173)};

\nextgroupplot[
    title={(f) 8B Novelty Rate $\uparrow$},
    ylabel={Fraction novel},
    ymin=0.1, ymax=0.45,
    xmin=1.6, xmax=5.4,
    xtick={2,3,4,5},
]
\addplot[cbase, dashed, mark=o, mark options={solid}] coordinates {(2,0.2117)(3,0.2264)(4,0.1814)(5,0.2235)};
\addplot[ctrain, mark=triangle*, mark options={solid}] coordinates {(2,0.2194)(3,0.1867)(4,0.1792)(5,0.2043)};
\addplot[cgen, mark=diamond*, mark options={solid}] coordinates {(2,0.3849)(3,0.3394)(4,0.3503)(5,0.3007)};
\addplot[cboth, mark=square*, mark options={solid}] coordinates {(2,0.3937)(3,0.366)(4,0.3424)(5,0.3331)};

\end{groupplot}

\path (ablplots c1r2.south west) -- (ablplots c3r2.south east) coordinate[midway, yshift=-0.8cm] (legendpos);
\node[anchor=north, font=\large] at (legendpos) {
    \tikz\draw[cbase, dashed, line width=1pt] (0,0) -- (0.6,0); \tikz\node[cbase,circle,draw,inner sep=0.9pt,line width=0.6pt] {}; Baseline \quad
    \tikz\draw[ctrain, line width=1pt] (0,0) -- (0.6,0); \tikz\node[ctrain,fill=ctrain,regular polygon,regular polygon sides=3,inner sep=0.8pt] {}; Train-only \quad
    \tikz\draw[cgen, line width=1pt] (0,0) -- (0.6,0); \tikz\node[cgen,fill=cgen,diamond,inner sep=0.8pt] {}; Gen-only \quad
    \tikz\draw[cboth, line width=1pt] (0,0) -- (0.6,0); \tikz\node[cboth,fill=cboth,rectangle,inner sep=1.2pt] {}; Both (VD75)
};

\end{tikzpicture}%
}
\caption{Embedding diversity by dropout phase ($\alpha{=}0.75$). Both phases combined achieves the highest diversity. All metrics use \texttt{text-embedding-3-small}.}
\label{fig:embedding_ablations}
\end{figure*}

%% file: tables/cross_scale.tex
\begin{table}[H]
\centering
\caption{Cross-scale co-evolution: 8B proposer $\to$ 4B solver, compared with symmetric 4B$\to$4B from \Cref{tab:vanilla_eval}. Best per group in \best{green bold}.}
\label{tab:cross_scale}
\resizebox{\columnwidth}{!}{%
\begin{tabular}{ll cccccc c}
\toprule
& & \multicolumn{6}{c}{\textbf{Pass@1 (\%)}} & \\
\cmidrule(lr){3-8}
\textbf{Proposer} & \textbf{Setting} & MATH500 & GSM8K & AMC & Olympiad & AIME'24 & AIME'25 & \textbf{Avg.} \\
\midrule
4B & Baseline & 64.2\tss{0.8} & \best{85.2}\tss{0.7} & 41.7\tss{1.4} & 23.2\tss{0.3} & 8.9\tss{0.9} & \best{6.7}\tss{1.6} & 38.3 \\
\rowcolor{oursrow} 4B & + VD85 & \best{65.7}\tss{0.9} & 81.8\tss{0.4} & \best{50.0}\tss{2.4} & \best{24.7}\tss{0.9} & \best{8.9}\tss{1.8} & 4.4\tss{0.9} & \best{39.3} \\
\midrule
8B & Baseline & 63.5\tss{2.7} & 82.7\tss{2.5} & \best{44.2}\tss{2.4} & \best{24.4}\tss{0.1} & 11.1\tss{1.6} & \best{6.7}\tss{5.4} & \best{38.8} \\
\rowcolor{oursrow} 8B & + VD85 (gen-only) & \best{66.1}\tss{1.1} & \best{83.4}\tss{2.7} & 35.8\tss{4.3} & 23.4\tss{0.7} & \best{13.3}\tss{2.7} & 2.2\tss{1.6} & 37.4 \\
\bottomrule
\end{tabular}%
}
\end{table}

%% file: tables/instruct.tex
\begin{table}[H]
\centering
\caption{Co-evolution on Qwen2.5-1.5B-Instruct (5 iterations).}
\label{tab:instruct}
\scriptsize
\setlength{\tabcolsep}{5pt}
\begin{tabular}{l cc}
\toprule
\textbf{Model} & MATH500 & AMC \\
\midrule
\textcolor{gray}{Base (no training)} & \textcolor{gray}{47.7} & \textcolor{gray}{25.3} \\
Baseline solver & 47.3 & \best{26.0} \\
\rowcolor{oursrow} VD75 solver & 47.4 & 25.2 \\
\bottomrule
\end{tabular}
\end{table}

%% file: figures/annealing.tex
\begin{figure}[!htb]
\centering
\begin{tikzpicture}

\definecolor{cbase}{RGB}{120,120,120}
\definecolor{cfixed}{RGB}{221,132,82}
\definecolor{canneal}{RGB}{31,138,112}

\pgfplotsset{
    anneal panel/.style={
        width=0.36\columnwidth,
        height=4.0cm,
        xmin=0.7, xmax=5.3,
        xtick={1,2,3,4,5},
        xlabel={Iteration},
        xlabel style={font=\small},
        ylabel style={font=\small},
        tick label style={font=\scriptsize},
        title style={font=\small\bfseries},
        grid=major,
        grid style={gray!15, dashed},
        every axis plot/.append style={line width=1.0pt, mark size=1.8pt},
        axis background/.style={fill=gray!4},
    },
}

\begin{groupplot}[
    group style={
        group size=3 by 1,
        horizontal sep=1.2cm,
        group name=annealplots
    },
    anneal panel,
]

\nextgroupplot[
    title={(a) Math Avg.},
    ylabel={Pass@1 (\%)},
    ymin=50.5, ymax=57.5,
    xticklabels={{\scriptsize 1\\\scriptsize\color{canneal!80!black}.75}, {\scriptsize 2\\\scriptsize\color{canneal!80!black}.81}, {\scriptsize 3\\\scriptsize\color{canneal!80!black}.88}, {\scriptsize 4\\\scriptsize\color{canneal!80!black}.94}, {\scriptsize 5\\\scriptsize\color{canneal!80!black}1.0}},
    xticklabel style={align=center},
]
\addplot[cbase, dashed, mark=o, mark options={solid}] coordinates {
    (1,53.7) (2,53.8) (3,53.5) (4,54.2) (5,54.2)
};
\addplot[cfixed, mark=square*, mark options={solid}] coordinates {
    (1,51.4) (2,54.8) (3,53.5) (4,54.2) (5,56.6)
};
\addplot[canneal, mark=triangle*, mark options={solid}] coordinates {
    (1,51.8) (2,54.7) (3,56.1) (4,55.1) (5,55.1)
};

\nextgroupplot[
    title={(b) MATH500},
    ylabel={Pass@1 (\%)},
    ymin=62, ymax=76,
    xticklabels={{\scriptsize 1\\\scriptsize\color{canneal!80!black}.75}, {\scriptsize 2\\\scriptsize\color{canneal!80!black}.81}, {\scriptsize 3\\\scriptsize\color{canneal!80!black}.88}, {\scriptsize 4\\\scriptsize\color{canneal!80!black}.94}, {\scriptsize 5\\\scriptsize\color{canneal!80!black}1.0}},
    xticklabel style={align=center},
]
\addplot[cbase, dashed, mark=o, mark options={solid}] coordinates {
    (1,66.9) (2,71.3) (3,66.7) (4,69.6) (5,69.5)
};
\addplot[cfixed, mark=square*, mark options={solid}] coordinates {
    (1,64.3) (2,71.1) (3,71.5) (4,68.8) (5,69.6)
};
\addplot[canneal, mark=triangle*, mark options={solid}] coordinates {
    (1,68.3) (2,70.6) (3,73.7) (4,73.2) (5,71.7)
};

\nextgroupplot[
    title={(c) OlympiadBench},
    ylabel={Pass@1 (\%)},
    ymin=20, ymax=30,
    xticklabels={{\scriptsize 1\\\scriptsize\color{canneal!80!black}.75}, {\scriptsize 2\\\scriptsize\color{canneal!80!black}.81}, {\scriptsize 3\\\scriptsize\color{canneal!80!black}.88}, {\scriptsize 4\\\scriptsize\color{canneal!80!black}.94}, {\scriptsize 5\\\scriptsize\color{canneal!80!black}1.0}},
    xticklabel style={align=center},
]
\addplot[cbase, dashed, mark=o, mark options={solid}] coordinates {
    (1,26.1) (2,26.1) (3,28.2) (4,27.5) (5,24.0)
};
\addplot[cfixed, mark=square*, mark options={solid}] coordinates {
    (1,24.8) (2,26.6) (3,22.2) (4,26.7) (5,28.6)
};
\addplot[canneal, mark=triangle*, mark options={solid}] coordinates {
    (1,25.3) (2,28.2) (3,27.1) (4,27.5) (5,28.3)
};

\end{groupplot}

\path (annealplots c1r1.south west) -- (annealplots c3r1.south east)
    coordinate[midway, yshift=-1.0cm] (legendpos);
\node[anchor=north, font=\small] at (legendpos) {
    \tikz\draw[cbase, dashed, line width=1pt] (0,0) -- (0.5,0); \tikz\node[cbase,circle,draw,inner sep=0.9pt,line width=0.6pt] {}; Baseline \quad
    \tikz\draw[cfixed, line width=1pt] (0,0) -- (0.5,0); \tikz\node[cfixed,fill=cfixed,rectangle,inner sep=1.1pt] {}; VD75 (fixed) \quad
    \tikz\draw[canneal, line width=1pt] (0,0) -- (0.5,0); \tikz\node[canneal,fill=canneal,regular polygon,regular polygon sides=3,inner sep=0.8pt] {}; VD75 (anneal)
};

\end{tikzpicture}
\caption{Qwen3-8B solver accuracy across iterations under fixed vs.\ annealed ($0.75 {\to} 1.0$) vocabulary dropout. Teal $\alpha$ values below each tick show the anneal schedule. (a) Mean of MATH500, GSM8K, OlympiadBench, and Minerva Math.}
\label{fig:annealing}
\end{figure}

%% file: tables/benchmarks.tex
\begin{table}[H]
\centering
\caption{Evaluation benchmarks.}
\label{tab:benchmarks}
\scriptsize
\setlength{\tabcolsep}{6pt}
\begin{tabular}{lrl}
\toprule
\textbf{Benchmark} & \textbf{\# Examples} & \textbf{Domain} \\
\midrule
GSM8K~\citep{cobbe2021gsm8k} & 1{,}319 & Grade-school math \\
MATH500~\citep{hendrycks2021math} & 500 & Competition math \\
AMC~\citep{amc2023} & 40 & Competition math \\
OlympiadBench~\citep{he2024olympiadbench} & 910 & Olympiad math \\
AIME 2024~\citep{aime2024} & 30 & Olympiad math \\
AIME 2025~\citep{aime2024} & 30 & Olympiad math \\
\midrule
\multicolumn{3}{l}{\textit{Additional benchmarks (annealing experiments)}} \\
Minerva Math~\citep{lewkowycz2022minerva} & 272 & STEM math \\
\bottomrule
\end{tabular}
\end{table}